\documentclass[lettersize,journal]{IEEEtran}
\usepackage{amsmath,amsfonts}
\usepackage{algorithmic}
\usepackage{algorithm}
\usepackage{array}
\usepackage{textcomp}
\usepackage{stfloats}
\usepackage{url}
\usepackage{verbatim}
\usepackage{graphicx}
\usepackage{cite}
\usepackage{color}
\usepackage{makecell}
\usepackage{subfigure}
\usepackage{multirow}
\usepackage{diagbox}
\usepackage{colortbl}
\usepackage{xcolor}
\usepackage{booktabs}
\usepackage{bm}
\usepackage[colorlinks,linkcolor=red,anchorcolor=blue,citecolor=green]{hyperref}
\definecolor{mygray}{gray}{.9}

\newtheorem{definition}{\textbf{Definition}}
\usepackage[normalem]{ulem} 
\newcommand\shadecolor{\bgroup\protect\markoverwith
  {\textcolor{mygray}{\rule[-.5ex]{2pt}{2.5ex}}}\protect\ULon}
\newcommand{\tabincell}[2]{
    \begin{tabular}{@{}#1@{}}#2\end{tabular}
}

\begin{document}

\title{Multi-Domain Active Learning: Literature Review and Comparative Study}

\author{
   Rui He, Shengcai Liu, Shan He, and Ke Tang
   \thanks{
      This work was supported in part by the Guangdong Provincial Key Laboratory under Grant 2020B121201001, in part by the Program for Guangdong Introducing Innovative and Entrepreneurial Teams under Grant 2017ZT07X386, and in part by the MOE University Scientific-Technological Innovation Plan Program.
   }
   \thanks{
      Rui He, Shengcai Liu and Ke Tang are with the Guangdong Provincial Key Laboratory of Brain-Inspired Intelligent Computation, Department of Computer Science and Engineering, Southern University of Science and Technology, Shenzhen 518055, China. E-mail: her2018@mail.sustech.edu.cn; liusc3@sustech.edu.cn, tangk3@sustech.edu.cn
   }
   \thanks{
      Rui He and Shan He are with the School of Computer Science, University of Birmingham, Birmingham, B15 2TT, United Kingdom.
      E-mail: s.he@cs.bham.ac.uk
   }
   \thanks{
      Corresponding author: Shengcai Liu.
   }
   \thanks{
      \textit{Manuscript accepted October 08, 2022, IEEE-TETCI. \copyright 2022 IEEE. Personal use of this material is permitted. Permission from IEEE must be obtained for all other uses, in any current or future media, including reprinting/republishing this material for advertising or promotional purposes, creating new collective works, for resale or redistribution to servers or lists, or reuse of any copyrighted component of this work in other works.}
   }
}

\markboth{Journal of \LaTeX\ Class Files,~Vol.~X, No.~X, MM~YYYY}%
{Shell \MakeLowercase{\textit{et al.}}: A Sample Article Using IEEEtran.cls for IEEE Journals}


\maketitle

\begin{abstract}
   Multi-domain learning (MDL) refers to learning a set of models simultaneously, where each model is specialized to perform a task in a particular domain.
   Generally, a high labeling effort is required in MDL, as data needs to be labeled by human experts for every domain.
   Active learning (AL) can be utilized in MDL to reduce the labeling effort by only using the most informative data.
   The resultant paradigm is termed multi-domain active learning (MDAL).
   In this work, we provide an exhaustive literature review for MDAL on the relevant fields, including AL, cross-domain information sharing schemes, and cross-domain instance evaluation approaches.
   It is found that the few studies which have been directly conducted on MDAL cannot serve as off-the-shelf solutions on more general MDAL tasks.
   To fill this gap, we construct a pipeline of MDAL and present a comprehensive comparative study of thirty different algorithms, which are established by combining six representative MDL models and five commonly used AL strategies.
   We evaluate the algorithms on six datasets involving textual and visual classification tasks.
   In most cases, AL brings notable improvements to MDL, and the naive BvSB (best vs. second best) Uncertainty strategy can perform competitively with the state-of-the-art AL strategies.
   Besides, BvSB with the MAN (multinomial adversarial networks) model can consistently achieve top or above-average performance on all the datasets.
   Furthermore, we qualitatively analyze the behaviors of the well-performed strategies and models, shedding light on their superior performance in the comparison.
   Finally, we recommend using BvSB with the MAN model in the application of MDAL due to their good performance in the experiments.
\end{abstract}

\begin{IEEEkeywords}
   Multi-domain learning, active learning, comparative study
\end{IEEEkeywords}

\section{Introduction}

\label{sec:introduction}

\IEEEPARstart{B}{uilding} classifiers on the data collected from different domains is common in real-world applications~\cite{MDL}.
Here, domains usually refer to different datasets under different distributions.
For example, in sentiment analysis of product reviews, distinct product categories are considered as different domains.
The conventional approach is to build one single model on all the domains jointly or build models on each domain independently.
However, the joint training eliminates the unique information of each domain, and the independent training neglects the correlation among domains.
Thus, both approaches usually bring suboptimal performance.
When the number of domains is large, such limitations can be more serious.
Under this circumstance, multi-domain learning (MDL) \cite{MDL} has been proposed to capture both the domain-invariant and the domain-specific information to overcome the aforementioned limitations.

In real life, high labeling effort is generally required in MDL, as data needs to be labeled by human experts for every domain.
Active learning (AL) \cite{active-learning-survey} is a general approach to reducing the labeling effort by interactively selecting and labeling the most informative instances in conventional single domain learning.
Hence, a natural question arises: can AL be used to reduce the labeling effort in MDL?
Despite the practical importance of this problem, only a few studies have been conducted on this issue.
The application of these studies is inherently limited as they are only tailored for ad hoc tasks on specific types of models.
Specifically, the strategy in \cite{MDAL-text} needs to evaluate the size of the model's version space \cite{version-space}, which is intractable for models other than SVMs.
Besides, the strategy in another work \cite{zhang2016multi} is specifically designed for Rating-Matrix Generative Model, which can only be used for the multi-domain recommendation problem.
Thus, it is unclear how AL would perform in MDL on more general tasks with more advanced neural networks.


To fill this gap, we study the problem of utilizing AL to reduce the high labeling effort in MDL, which is termed multi-domain active learning (MDAL) \cite{MDAL-text}.
First, we provide an exhaustive literature review on the relevant fields that can serve as the foundations of MDAL, including pool-based AL, cross-domain information sharing schemes, and cross-domain instance evaluation approaches.
Then, we construct an easy-to-implement pipeline of MDAL, which can utilize the modern neural network--based MDL models and the conventional AL strategies.
Subsequently, a comprehensive comparative study of thirty different MDAL algorithms is presented.
The algorithms are established by combining six representative MDL models and five commonly used AL strategies.
We evaluate the algorithms on six datasets involving textual and visual classification tasks.
Besides, the well-performed models and strategies are qualitatively analyzed for their superiority.
The implementation\footnote{https://github.com/SupeRuier/mdal-pipeline} is available online so that people can easily adopt the pipeline to their MDAL tasks.
The main contributions can be summarized as follows:
\begin{itemize}
   \item We provide the first literature review and the formal definition for MDAL.
         In addition, a pipeline is constructed as an off-the-shelf solution for MDAL.
   \item To the best of our knowledge, this is the first comparative work for MDAL.
         The results show that the naive BvSB (best vs. second best) Uncertainty strategy could perform competitively with the state-of-the-art AL strategies.
         Besides, BvSB with the MAN (multinomial adversarial networks) model can consistently achieve top or above-average performance on all the datasets.
         Thus, we recommend using BvSB with the MAN model in the application of MDAL.
   \item Through further investigations, we claim that the shared-private structure is responsible for the superiority of the MAN model, and the high intra-batch diversity is associated with the superiority of Uncertainty.
         Moreover, we also point out several future directions in MDAL.
\end{itemize}

The remainder of this article is organized as follows.
Section~\ref{sec:definitions} provides the formal definitions of AL, pool-based AL, MDL and MDAL.
Section~\ref{sec:review} presents the literature review on the relevant research fields which serve as foundations for MDAL.
The details of the MDAL pipeline are described in Section~\ref{sec:pipeline}.
Section~\ref{sec:comparison} presents settings of the comparative study.
The comparison results are shown in Section~\ref{sec:results}.
The outstanding models and strategies in the comparison are further investigated in Section~\ref{sec:investigations}.
Section~\ref{sec:conclusion} concludes the article with discussions and future directions.

\section{Problem Definitions}
\label{sec:definitions}

In this section, we first provide the definition of active learning.
Specifically, \textit{pool-based active learning} is mainly considered because it is the most basic AL scenario and the most directly related AL scenario to MDAL.
Next, the definitions of \textit{multi-domain learning} and \textit{multi-domain active learning} are provided.

\begin{definition}[Active learning] \label{def:raw-al}
   AL \cite{active-learning-survey} is to reduce the labeling cost by only annotating informative instances rather than all the unlabeled data.
   The informative labeled set $\mathcal{L_\alpha}$ is selected from the unlabeled data $\mathcal{U}$ by an AL acquisition function (strategy) $\alpha$.
   In the meantime, a model $\mathcal{M}_\alpha$ can be trained by using the selected informative data.
   The unlabeled data $\mathcal{U}$ can be pre-collected as a data pool (pool-based AL) or come in a steam manner (steam-based AL).
\end{definition}

\begin{definition}[Pool-based active learning] \label{def:al}
   Pool-based AL \cite{DAL-survey} is to select informative instances from the unlabeled data pool $\mathcal{U}_0$ which has been obtained at the very beginning.
   The instances are iteratively selected according to an AL acquisition function $\alpha$.
   First, a base model $\mathcal{M}_0$ is trained on the initial labeled data $\mathcal{L}_0$.
   Then, a batch of to-be-queried instances $\mathcal{Q}_i$ with top-$b$ acquisition scores is selected and annotated by an oracle in the $i$-th AL iteration:
   \begin{equation}
      \mathcal{Q}_i = \arg \max_{x \in \mathcal{U}_{i-1}}^{b} \alpha (x, \mathcal{M}_{i-1}), \quad |\mathcal{Q}_i| = b
   \end{equation}
   $\mathcal{L}_{i-1}$ and $\mathcal{U}_{i-1}$ are then updated with the selected batch $\mathcal{Q}_i$.
   In the meantime, the model $\mathcal{M}_{i}$ is trained on the updated data $\mathcal{L}_{i}$ and $\mathcal{U}_{i}$.
   The labeling process terminates once the labeling budget $\mathcal{B}$ is exhausted or the desired performance has been reached.
   Finally, the labeled set $\mathcal{L}_{i}$ and the model $\mathcal{M}_{i}$ at the final iteration are obtained as the outputs.
\end{definition}

\begin{definition}[Multi–domain learning] \label{def:mdl}
   Given $K$ different data sources (domains) $\mathcal{D} = \{\mathcal{D}_1, $ $ \mathcal{D}_2, \dots, \mathcal{D}_K\}$, a set of data pools $\mathcal{P} = \{\mathcal{P}_1, \mathcal{P}_2,\dots, \mathcal{P}_K\}$ containing both labeled and unlabeled data is collected from $\mathcal{D}$ in advance.
   The labeled data from each pool constitute a labeled data set $\mathcal{L} = \{\mathcal{L}_1, $ $ \mathcal{L}_2, \dots, \mathcal{L}_K\}$.
   MDL is to find a set of models $\mathcal{M} = \{\mathcal{M}_1,\mathcal{M}_2,\dots,\mathcal{M}_K\}$ for $K$ domains by utilizing the common knowledge of different domains, which can be expressed as follows:
   \begin{equation}
      \begin{aligned}
         \min_{\mathcal{M}} {Loss}_{\rm sup}(\mathcal{M}; \mathcal{L}) + \Omega (\mathcal{M}; \mathcal{P})
      \end{aligned}
   \end{equation}
   ${Loss}_{\rm sup}(\mathcal{M}; \mathcal{L})$ denotes the supervised loss on the labeled set $\mathcal{L}$.
   $\Omega(\mathcal{M}; \mathcal{P})$ denotes a designed loss on the set of data pools $\mathcal{P}$ for capturing the common knowledge through model $\mathcal{M}$.
\end{definition}

\begin{definition}[Multi–domain active learning] \label{def:mdal}
   MDAL is to reduce the labeling cost in MDL by only selecting informative instances from $K$ different domains $\mathcal{D} = \{\mathcal{D}_1, $ $ \mathcal{D}_2, \dots, \mathcal{D}_K\}$.
   First, a set of base models $\mathcal{M} = \{\mathcal{M}_1,\mathcal{M}_2,\dots,\mathcal{M}_K\}$ is trained on a small initial set of labeled data $\mathcal{L} = \{\mathcal{L}_1, $ $ \mathcal{L}_2, \dots, \mathcal{L}_K\}$ from the pre-collected set of data pools $\mathcal{P}$.
   Then, in each iteration, a set of to-be-queried data $\mathcal{Q} = \{\mathcal{Q}_1, $ $ \mathcal{Q}_2, \dots, \mathcal{Q}_K\}$ from $\mathcal{P}$ is selected and annotated according to an acquisition function $\alpha$.
   After the selection, $\mathcal{L}$ updates, and the set of models $\mathcal{M}$ is retrained on the updated $\mathcal{L}$ and $\mathcal{P}$ in the meantime.
   The labeling process terminates once the labeling budget $\mathcal{B}$ is exhausted or the desired performance is reached.
   Finally, the labeled set $\mathcal{L}$ and the set of models $\mathcal{M}$ at the final iteration are obtained as the outputs.
\end{definition}

\section{Literature Review}
\label{sec:review}

\begin{figure}[htbp]
   \centerline{\includegraphics[width=\linewidth]{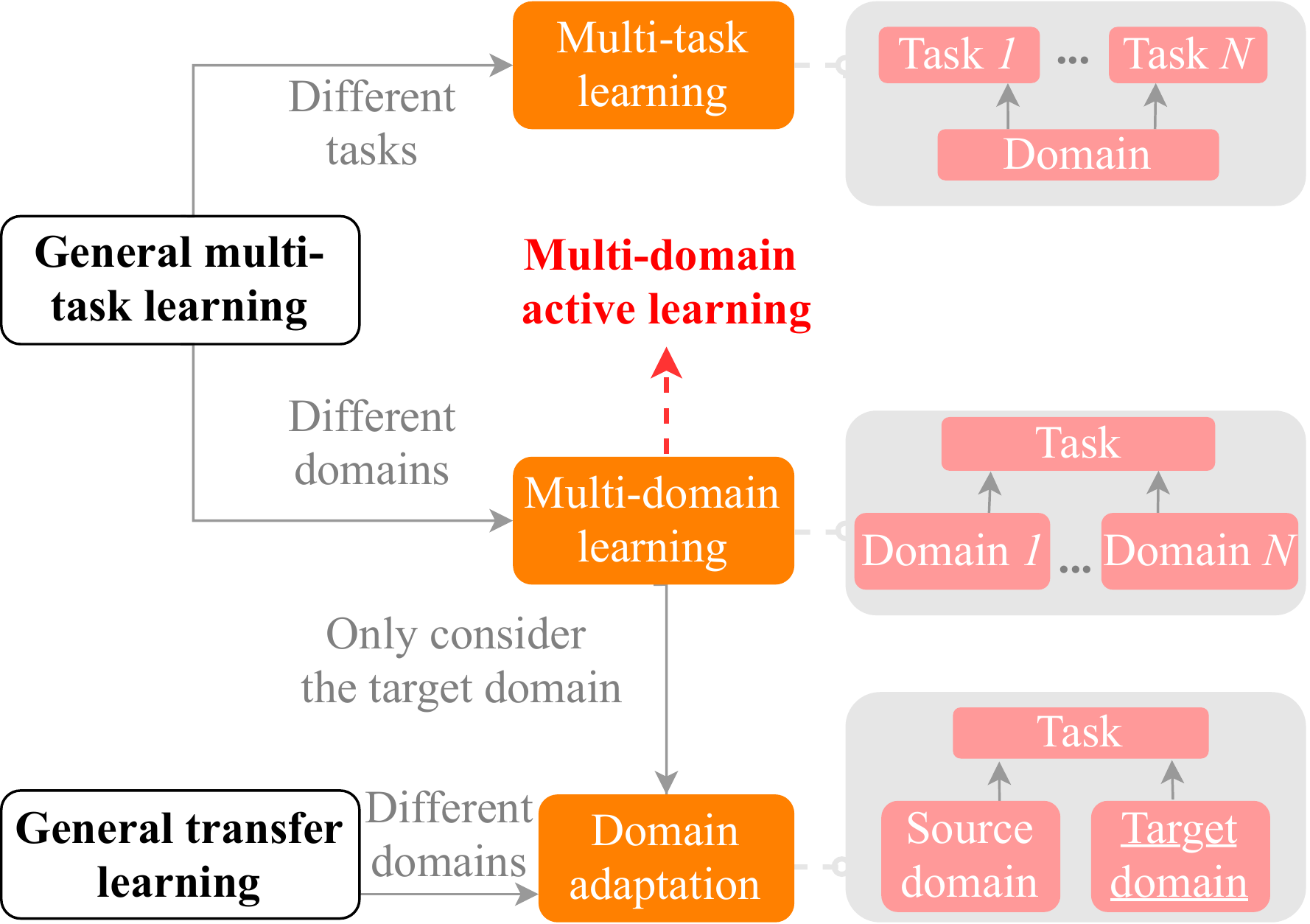}}
   \caption{The relations among different terminologies. The sub-figures with gray backgrounds are sketches of problem formulations.}
   \label{fig:field-relations}
\end{figure}

To better understand MDAL, first, some relevant fields such as multi-task learning (MTL) \cite{mtl-survey} and domain adaptation (DA) \cite{TL-survey} are distinguished in Fig.~\ref{fig:field-relations}.
Although MDL is sometimes referred to as a type of (general) multi-task learning, it usually focuses on the same task on multiple domains.
However, narrowly defined MTL focuses on different tasks on the same domain.
Differently, DA is a subfield of transfer learning, where the knowledge is transferred from one domain to another.
DA concentrates on the target domain's performance, while MDAL focuses on all domains' performance.

For example, with the annotated shopping review data, the reviews of the same product can be considered as one domain.
The reviews can be used to train models for different tasks, such as sentiment classification and named entity recognition tasks.
MTL can simultaneously build models for multiple tasks on one domain (reviews of one product).
In contrast, MDL builds models on multiple domains (multiple products' reviews) but only for one task, e.g., sentiment classification task.
Similar to MDL, DA also focuses on one task.
However, DA focuses on building a model for another target domain (reviews of another product) by using the source domain data (annotated reviews of the current product).

This section reviews a relatively broad set of fields that serve as foundations of MDAL.
As an intersection of AL and MDL, MDAL evaluates instances to select the most valuable ones for training and utilizes information-sharing schemes to build models.
Thus, the instance evaluation approaches and the practical considerations in pool-based AL are reviewed in Section~\ref{sec:active-learning}.
Besides, Section~\ref{sec:multiple-domains} reviews cross-domain information-sharing schemes of both domain adaptation and multi-domain learning to guide the model training.
Finally, cross-domain instance evaluation approaches from both active domain adaptation and multi-domain active learning are reviewed in Section~\ref{sec:active-learning-domains}, as the evaluation approaches can be different from simple pool-based AL with the existence of multiple domains.

\subsection{Pool-Based Active Learning}
\label{sec:active-learning}

Pool-based AL aims to obtain a subset of informative labeled instances to maximize the task model's performance.
Several surveys \cite{AL-Comparative-Survey, DAL-survey} have reviewed the recent advances of AL.
We also maintain an open-sourced AL knowledge base, named \textit{awesome active learning}\footnote{https://github.com/SupeRuier/awesome-active-learning}, tracking the latest AL research in both technical developments and applications.
The primary taxonomy and the representative works of the current AL strategies are shown in Table~\ref{table:al-taxonomy}.
The taxonomy has two dimensions: \textit{evaluation criteria} (Section~\ref{sec:info-based} to \ref{sec:hybrid}) and \textit{practical considerations} (Section~\ref{sec:batch} \& \ref{sec:advanced}).
The evaluation criteria explain how samples are selected for labeling.
Practical considerations are relevant to the actual requirements encountered in real applications.

\begin{table*}[htbp]
   \caption{The taxonomy for pool-based AL and the representative works.}
   \centering
   \begin{tabular}{cll}
      \toprule
      \textbf{Evaluation Criteria}                                                      & \textbf{Sub-Categories}        & \textbf{Representative Works}                                                                  \\
      \midrule
      \multirow{3}[3]{*}{Informativeness-based (Sec.~\ref{sec:info-based})}             & Uncertainty Sampling           & LC \cite{Uncertainty}, Margin (BvSB) \cite{BvSB}, DFAL \cite{DFAL}, CAL \cite{CAL}             \\
      \cmidrule(l){2-3}
                                                                                        & Disagreement Sampling          & QBC \cite{QBC}, BALD \cite{BALD}, MC-dropout \cite{MCdropout}, MeanSTD \cite{meanstd}          \\
      \cmidrule(l){2-3}
                                                                                        & Expected Improvements Sampling & EGL \cite{EGL}, MOCU \cite{MOCU}                                                               \\
      \midrule
      \multirow{3}[3]{*}{Representativeness-impart (Sec.~\ref{sec:representativeness})} & Cluster-based sampling         & BADGE \cite{BADGE}, Coreset \cite{coreset}                                                     \\
      \cmidrule(l){2-3}
                                                                                        & Density-based sampling         & Information density \cite{EGL}                                                                 \\
      \cmidrule(l){2-3}
                                                                                        & Alignment-based sampling       & WAAL \cite{WAAL}, VAAL \cite{VAAL}, DAL \cite{DAL}, SCAL \cite{SCAL}                           \\
      \midrule
      Learn to score (Sec.~\ref{sec:learn2score})                                       & ------                         & ALBL \cite{ALBL}, LAL \cite{LAL}                                                               \\
      \midrule
      Others (Sec.~\ref{sec:hybrid})                                                    & ------                         & SPAL \cite{SPAL}, SPMCAL \cite{SPAL-cnn}                                                       \\
      \bottomrule
      \toprule
      \textbf{Considerations}                                                           & \textbf{Sub-Categories}        & \textbf{Representative Works}                                                                  \\
      \midrule
      Batch mode (Sec.~\ref{sec:batch})                                                 & ------                         & BADGE \cite{BADGE}, BatchBALD \cite{BatchBALD}, Active-DPP \cite{active-dpp}, BAIT \cite{BAIT} \\
      \midrule
      \multirow{2}[3]{*}{Beyond strategies (Sec.~\ref{sec:advanced})}                   & Data                           & BGADL \cite{BGADL}, LADA \cite{LADA}                                                           \\
      \cmidrule(l){2-3}
                                                                                        & Training procedure             & CEAL \cite{CEAL}                                                                               \\
      \bottomrule
   \end{tabular}
   \label{table:al-taxonomy}
\end{table*}

\subsubsection{Informativeness-Based}
\label{sec:info-based}

Informativeness-based methods define how much the current inference system is uncertain about the output of the corresponding instance.

\textbf{Uncertainty Sampling} \cite{Uncertainty} uses the probability output to evaluate the uncertainty of the example.
Specifically, the most likely possibility (Least Confident, or LC), the margin between the top-2 classes (Margin or BvSB \cite{BvSB}), and the output entropy (Entropy) can be used to evaluate instances.
DeepFool active learning (DFAL) \cite{DFAL} proposes to use the distances between the examples and their adversarial examples as the evaluations since measuring the exact distance to the decision boundaries is intractable.
Contrastive active leaning (CAL) \cite{CAL} considers the inconsistency of predictions with the neighbors as the selection criteria.

\textbf{Disagreement Sampling} uses the disagreement of the outputs from different models as the informativeness evaluation when a bag of models is available in the inference system.
Query-by-committee (QBC) \cite{QBC} utilizes the disagreement of multiple classifiers.
Bayesian active learning by disagreement (BALD) \cite{BALD} is based on a probabilistic Gaussian process classifier.
It seeks instances in which the model parameters under the posterior disagree about the outcome the most.
Monte-Carlo (MC) dropout \cite{MCdropout} extends BALD to a Bayesian CNN model, and the prediction uncertainty is induced by marginalizing over the approximate posterior using Monte Carlo integration (models sampled with dropout).
Mean standard deviation (MeanSTD) \cite{meanstd} computes the standard deviation over the softmax outputs of Monte Carlo samples as an uncertainty evaluation.

\textbf{Expected Improvements Sampling} considers the instance that most improves the model's performance as the informative instance.
This term is usually heuristically calculated by how much the instance can influence the model, for example, the expected gradient length (EGL) \cite{EGL,EGL_text}.
Several works maximize the expected loss reduction \cite{ELR} or the expected variance reduction \cite{EVR} in a one-step-look-ahead manner.
Some methods aim to reduce the model uncertainty related to the classification error, such as mean objective cost of uncertainty (MOCU) \cite{MOCU}.

\subsubsection{Representativeness-Impart}
\label{sec:representativeness}

The informativeness-based strategies usually focus on the decision boundary and neglect the data distribution.
In this situation, many works take the representativeness of the data into account.
Representativeness measures how much the labeled instances are aligned with the unlabeled instances in distribution.
We note that representativeness is commonly used with informativeness for sampling.

\textbf{Cluster-based sampling} evaluates instances in a pairwise-matrix form, and no scores are provided for the selection.
As a two-stage approach, a pre-selection or a data preprocessing procedure needs to be carried out in advance.
Then, the selection can be conducted on the pre-selected informative instances \cite{svm-cluster, cluster-margin, DBAL} or the induced gradients such as batch active learning by diverse gradient embeddings (BADGE) \cite{BADGE}.
The clustering can also be applied to the original features for further informativeness selection \cite{pre-cluster}.
Without considering informativeness, Coreset \cite{coreset} formulates AL as a core-set selection.
The model trained on the selected core set can perform as closely as possible to the model trained on the entire dataset.

\textbf{Density-based} and \textbf{Alignment-based sampling} provide scores as informativeness-based methods do.
Thus, the acquisition function can easily be calculated as
$\alpha_{\text{overall}} = \alpha_{\text{informativeness}} + \lambda \times \alpha_{\text{representativeness}}$, where the $\lambda$ is a trade-off parameter.
Density-based sampling points out that the location with more density should be more likely to be queried, e.g., information density \cite{EGL}.
Alignment-based sampling directly considers the distribution alignment between labeled (selected) data and unlabeled data.
Discriminative active learning (DAL) \cite{DAL}, variational adversarial active learning (VAAL) \cite{VAAL}, wasserstein adversarial AL (WAAL) \cite{WAAL}, dual adversarial network (DAAL) \cite{DAAL} and MinimaxAL \cite{MinimaxAL} attempt to utilize the distinguishability to evaluate the representativeness.
The instances with a higher probability of being unlabeled data will be selected under the evaluation of the discriminator.
Supervised contrastive active learning (SCAL) \cite{SCAL} selects the least similar instances to the labeled ones with the same class in the embedding space.

\subsubsection{Learn to Score}
\label{sec:learn2score}

It is hard to find a single strategy that dominates all the learning tasks.
Several works try to learn an acquisition function from the labeling process instead of manually designing the function.
Active learning by learning (ALBL) \cite{ALBL} utilizes multiple strategies as a multi-armed bandit learner and estimates the performance of different strategies on the fly.
Learning active learning (LAL) \cite{LAL} trains a random forest regressor to predict the expected error reduction for a candidate sample at particular learning states.
Learning loss for active learning (LLAL) \cite{LLAL} attaches a small parametric ``loss prediction module'' to a target network for predicting target losses of unlabeled inputs.
Some other works formulate the AL process as a Markov decision process and try to solve it as a reinforcement learning \cite{ALRL} or an imitation learning \cite{ALIL} problem.
Learn from historical sequences (LHS) \cite{LHS} is to learn an active learning ranker from the previous samples to guide the selection.

\subsubsection{Others}
\label{sec:hybrid}

Except informativeness and representativeness, other innovative heuristics can also be involved.
Considering the easiness of instances, SPAL (self-paced active learning) \cite{SPAL} and SPMCAL (self-paced multi-criteria active learning) \cite{SPAL-cnn} consider AL as a self-paced learning problem and select from the easy instances to the hard ones.

\subsubsection{Batch AL}
\label{sec:batch}

Examples are queried in batches in the practical training process to ensure effective training.
Conventional active learning tends to greedily select instances with the top evaluation scores from the acquisition function, leading to information overlap.
Batch-mode AL (also called diversity-based AL) tries to solve this myopic problem by diversifying the selected batch.
This type of method has huge overlaps with representativeness-impart methods, which unintentionally increase the diversity of the selected batch, such as the two-stage strategies introduced in Section~\ref{sec:representativeness}, e.g., BADGE \cite{BADGE}.
The diversity can also be intentionally ensured by using greedy selection such as K-center-greedy \cite{coreset}, BatchBALD \cite{BatchBALD}, and BAIT (batch active learning via information matrices) \cite{BAIT}.
Active-DPP \cite{active-dpp} utilizes Determinantal Point Processes to diversify the batch.

\subsubsection{Beyond strategies}
\label{sec:advanced}

There are three main aspects of an AL process: a strategy, data, and a model under a specific training paradigm.
Conventionally, the model is trained in a fully supervised way, without any additional manipulation.
In this case, the improvements for the whole system should solely come from the AL strategy.
Considering the other two aspects, the performance of the whole system can be further improved.

In the aspect of \textit{data}, the data augmentation \cite{data-aug} approaches have been widely used in supervised learning.
The augmented data can be used in model training to enlarge the labeled training set, e.g., Bayesian generative active deep learning (BGADL) \cite{BGADL}.
With data augmentation, look-ahead data acquisition (LADA) \cite{LADA} further takes the influence of the augmented data into account to construct the acquisition function.

In the aspect of \textit{training procedure}, several works combine AL with semi-supervised learning (SemiSL) \cite{semiSL}.
The unlabeled data can be utilized to train the corresponding model.
Inspired by the wrapper methods in SemiSL, several works consider the pseudo-labels in the model training \cite{rethinkingAL}, such as cost-effective active learning (CEAL) \cite{CEAL}.
Several other semi-supervised AL works use data perturbation \cite{SSAL_consistancy} to increase the robustness.
The distinguishability of the labeled and unlabeled data can also be used to train the classifier \cite{TJLS}.
Some other works utilize incrementally training \cite{SCAL}, instead of retraining the whole model at each AL iteration.
The pretraining scheme is also used in AL to increase the performance \cite{pretrainAL, SelfSL-meets-AL}.

\subsection{Cross-domain information-sharing schemes}
\label{sec:multiple-domains}

This section reviews the multi-domain information sharing schemes in the relevant research fields, such as domain adaptation \cite{TL-survey}, multi-domain learning, and domain generalization \cite{DG}.
The information-sharing usually assumes that the knowledge from one domain can also assist the learning in another domain.
The relevant approaches are classified by the carrier of the information.
The primary taxonomy and the representative works of the current cross-domain information-sharing schemes are shown in Table~\ref{table:mdl-taxonomy}.

\begin{table*}[htbp]
   \caption{The taxonomy for cross-domain information-sharing schemes and the representative works.}
   \centering
   \begin{tabular}{cll}
      \toprule
      \textbf{Schemes}                                          & \textbf{Sub-Categories}                & \textbf{Representative Works}                                                                     \\
      \midrule
      Sharing instances (Sec.~\ref{sec:share-instance})         & ------                                 & TrAdaBoost \cite{TrAdaBoost}                                                                      \\
      \midrule
      \multirow{4}[3]{*}{\tabincell{c}{Sharing domain-invariant                                                                                                                                              \\representations through models\\ (Sec.~\ref{sec:share-representation})}} & Invariant representations only         &  TCA \cite{TCA}, DAN \cite{DAN}, DANN \cite{DANN}, AADA \cite{ADDA}                    \\
      \cmidrule(l){2-3}
                                                                & With domain-specific representations   & FEDA \cite{FEDA}, DSN \cite{DSN}, MAN \cite{MAN}, CAN \cite{CAN}                                  \\
      \cmidrule(l){2-3}
                                                                & With domain-specific classifiers       & MDNet \cite{MDNet}                                                                                \\
      \cmidrule(l){2-3}
                                                                & With other domain-specific information & Domain-guided dropout \cite{xiao2016learning}                                                     \\
      \midrule
      Sharing model components (Sec.~\ref{sec:share-component}) & ------                                 & Residual adapter \cite{rebuffi2017learning, rebuffi2018efficient}, CovNorm \cite{li2019efficient} \\
      \bottomrule
   \end{tabular}
   \label{table:mdl-taxonomy}
\end{table*}

\subsubsection{Sharing instances across domains}
\label{sec:share-instance}
Directly utilizing the instances from one domain to train a model for another domain is referred to as instance transfer in domain adaptation \cite{TL-survey}.
The essence is that some instances can be reused for another domain as they do not contain misleading information.
The data from the source domain can be directly used to train the target classifier as auxiliary examples \cite{direct-DA}.
Further, Jiang and Zhai \cite{discard-DA} removed the misleading instances to train the target model.
Instead, transfer adaBoost (TrAdaBoost) \cite{TrAdaBoost} sets weights to source data to balance their contribution to the target domain.

\subsubsection{Sharing domain-invariant representations through models}
\label{sec:through-model}
Distributions from different domains can be unified to eliminate the negative effects of the domain-specific features.
A domain-invariant representation can be learned for the subsequent inferences.

\textbf{Sharing domain-invariant representations only:}
\label{sec:share-representation}
Plenty of works match the marginal distribution by minimizing a discrepancy loss, such as the Maximum Mean Discrepancy (MMD).
Transfer component analysis (TCA) \cite{TCA} is a classic method that applies this idea to conventional kernel-based models.
Deep adaptation networks (DAN) \cite{DAN} and joint adaptation networks (JAN) \cite{JAN} adopt this idea to neural network-based domain adaptation models.
Several works utilize adversarial training to match the marginal distributions.
The representation extraction can be effective if a discriminator cannot tell which domain the instances come from.
Domain-adversarial neural networks (DANN) \cite{DANN}, adversarial discriminative domain adaptation (ADDA) \cite{ADDA}, and conditional adversarial domain adaptation (CADA) \cite{CADA} add discriminators into their models for domain adaptation.
Feng \textit{et al.} \cite{feng2019self} also adapted this idea to learn representations in MDL.

Some other works proposed to match conditional distributions.
The instances in the same class from different domains are expected to be projected closer.
Deep supervised domain adaptation (SDA) \cite{Deep_SDA}, multi-adversarial domain adaptation (MADA) \cite{MADA}, and transferrable prototypical networks (TPN) \cite{pan2019transferrable} match the distributions either by the existing true labels or the created pseudo-labels on the target domain for domain adaptation.
Saito \textit{et al.} \cite{saito2018maximum} tried to align the distributions by matching the decision boundaries.

\textbf{Sharing domain-invariant representations with domain-specific information:}
Solely learning the domain-invariant representations wastes the unique information of each domain from the data.
Thus, domain-specific information can further be included to guide the inference.

The domain-invariant and domain-specific representations can be concatenated together to make predictions.
Frustratingly easy domain adaptation (FEDA) \cite{FEDA} first applies this idea to MDL with linear models.
Domain separation networks (DSN) \cite{DSN} and MAN \cite{MAN} utilize the concatenation into neural networks using shared and private feature extractors.
Conditional adversarial networks (CAN) \cite{CAN} further include the conditional distribution matching in training the shared extractor.

Besides private feature extractors, domain-specific information can also be handled in other forms.
Multi-domain convolutional neural networks (MDNet) \cite{MDNet} apply different domain-specific classifiers to the shared representations for the multi-domain visual tracking problem.
Similarly, Xiao \textit{et al.} \cite{xiao2016learning} proposed a domain-guided dropout for a multi-domain person re-identification problem.
Besides, when there is more than one feature extractor (or channel), the representations \cite{li2019semi} or channels \cite{xiao2020multi} can be weighted differently to make predictions on different domains.

\subsubsection{Sharing model components}
\label{sec:share-component}
Instead of explicitly sharing the same feature extractor across domains for representation learning, the modularized model components can be shared for transferring the domain information.
Some works \cite{rebuffi2017learning, rebuffi2018efficient} design networks with domain-specific residual adapters and remain the rest components the same.
Li and Vasconcelos \cite{li2019efficient} involved a domain-specific CovNorm layer instead.

\subsection{Cross-Domain Instance Evaluation for AL}
\label{sec:active-learning-domains}

Selecting instances from different domains for AL requires proper cross-domain instance evaluation criteria.
The instances most beneficial to a certain target domain (active domain adaptation) or all the domains (multi-domain active learning) will be selected.
The works in this field are usually strongly related to the models introduced in Section~\ref{sec:through-model}, as the AL strategies are usually model-dependent.

\subsubsection{Active domain adaptation}
The AL strategies for active domain adaptation usually select the instances that can minimize the domain discrepancy to learn the domain-invariant representations better.
Chattopadhyay \textit{et al.} \cite{chattopadhyay2013joint} tried to select instances from the target domain by reducing MMD between the labeled and the unlabeled data.
Huang and Chen \cite{huang2016transfer} utilized this idea in another problem setting, where the selection was from the source domain.
Su \textit{et al.} \cite{su2020active} proposed active adversarial domain adaptation (AADA), where a discriminator is trained to weight the uncertainty scores of the target unlabeled examples.

\subsubsection{Multi-domain active learning}
Only a few works are directly related to MDAL, where the domain-shared representations are used with the domain-specific representations.
Li \textit{et al.} \cite{MDAL-text} proposed to use multiple SVM models on the concatenated features for each domain in multi-domain sentiment classification.
The instances that can mostly reduce the version space of the SVM models are selected.
Zhang \textit{et al.} \cite{zhang2016multi} selected the user-item pairs leading to the lowest global generalization error of the model in the application of multi-domain recommendation.
The application of these studies is inherently limited as they are only tailored for ad hoc tasks on specific types of models.
Hence, these approaches cannot be adapted to more advanced neural networks.

\section{The Pipeline for MDAL}
\label{sec:pipeline}

\begin{figure}[tbp]
   \centering
   \includegraphics[width=\linewidth]{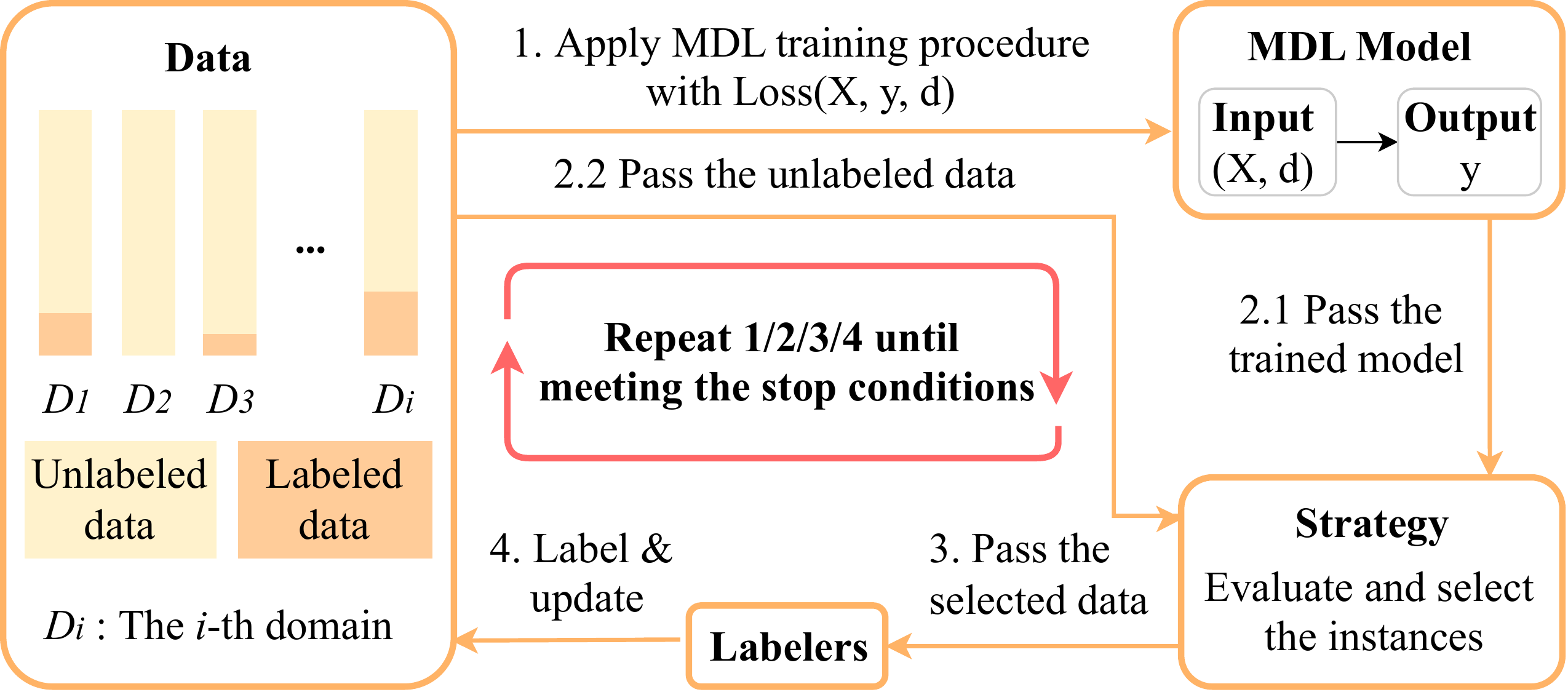}
   \caption{The proposed MDAL pipeline combines the MDL models and the conventional AL strategies.}
   \label{fig:mdal-pipeline}
\end{figure}

We propose a pipeline for MDAL, as shown in Fig.~\ref{fig:mdal-pipeline}.
Compared to the conventional single-domain AL, the differences are in data formulations, model structures and model training procedures.
Combining the models for MDL and the conventional AL strategies within this pipeline can be utilized as an off-the-shelf solution for MDAL.
The combination makes this pipeline easy to be implemented and adaptable to many alternative models and strategies.

Conventional AL strategies can be used in our MDAL pipeline to select the instances from different domains.
Specifically,
(1) the score-based strategies calculate a score for each instance as the evaluation.
The scores from different domains are gathered together and ranked, and the instances with the top-K scores are selected.
(2) The two-stage strategies select instances without calculating their scores.
Instead, they obtain the gradients or the embeddings of the instances as evaluations in the first stage.
Then, the evaluations from different domains are gathered together as a unified pool.
The strategies select instances from the pool in the second stage, as they do in conventional single domain AL.

With this pipeline, we are curious about:
\begin{enumerate}
   \item Is there a model or an information-sharing scheme naturally suitable for MDAL?
   \item Would AL bring improvements over the random selection? Is there any strategy that significantly outperforms the others?
   \item Would the models and strategies which obtain good overall performances still perform consistently well in each domain?
\end{enumerate}

A series of comparisons should be conducted to answer these questions.
Different models and strategies need to be compared on various datasets in the pipeline.

\section{Design of Comparative Experiments}
\label{sec:comparison}

This section describes the setup of the comparison experiments.
The selected datasets, models, and AL strategies are introduced in Section~\ref{sec:datasets}, \ref{sec:compared-models} and \ref{sec:compared-strategies}, respectively.
Section~\ref{sec:implementation} describes the details of implementations for model training and AL procedure.
The performance metrics are introduced in Section~\ref{sec:metrics}.

\subsection{Datasets}
\label{sec:datasets}

Six datasets are selected from the literature on MDL and domain adaptation problems.
These datasets at least contain two domains.
All the tasks are classification tasks, and the categories are the same across domains.
The details of the train-validation-test split can be found in the supplementary materials.

\begin{itemize}
   \item \textbf{Amazon}
         contains four domains: books, dvd, electronics, and kitchen.
         The raw sentences are processed by marginalized denoising autoencoders (mSDA) \cite{mSDA}.
   \item \textbf{Office-31}
         \cite{office31} contains thirty-one categories from three domains: Amazon, Webcam, and DSLR.
         Briefly, the DeCaf representations \cite{Decaf} are used.
   \item \textbf{Office-Home}
         \cite{office-home} contains four distinct domains: Art, Clip-Art, Product, and Real-world.
         Each of the four domains has sixty-five categories.
   \item \textbf{ImageCLEF}\footnote{\url{https://www.imageclef.org/2014/adaptation}}
         is a well-balanced dataset containing twelve categories from three domains: caltech, pascal, and imagenet.
   \item \textbf{Digits}
         is used in \cite{DANN}, which contains two domains (sub-datasets): MNIST \cite{mnist} and MNIST-M.
         MNIST-M is generalized by blending digits from the MNIST dataset over patches randomly extracted from color photos from BSDS500 \cite{bsds500}.
   \item \textbf{PACs}
         \cite{PACs} is an image dataset that contains seven categories from four domains: art-painting, cartoon, photo, and sketch.
\end{itemize}

\subsection{Models}
\label{sec:compared-models}

Most MDL models transfer information across domains by sharing domain-invariant representations, as we introduced in Section~\ref{sec:share-representation}.
Four models are selected for the comparison: \textbf{DANN} \cite{DANN}, \textbf{MDNet} \cite{MDNet}, \textbf{MAN} \cite{MAN}, and \textbf{CAN} \cite{CAN}.
The selected models are very representative of how shared and private information is handled, and they cover most of the current information-sharing schemes.
In addition, these models or the schemes behind them do not stick to the network structures in their original paper.
Instead, they could be adapted to other backbone neural networks.
Thus, it is convenient to implement these models on different types of tasks and datasets.

Two baseline models are also included: \textbf{SDL-separate} and \textbf{SDL-joint}.
There is no specifically designed information sharing schemes in these two models.
In SDL-separate, multiple networks are independently trained on the corresponding domains.
In SDL-joint, a single neural network is trained on the mixed data from all the domains.

\begin{figure*}[htbp]
   \centering
   \includegraphics[width=\linewidth]{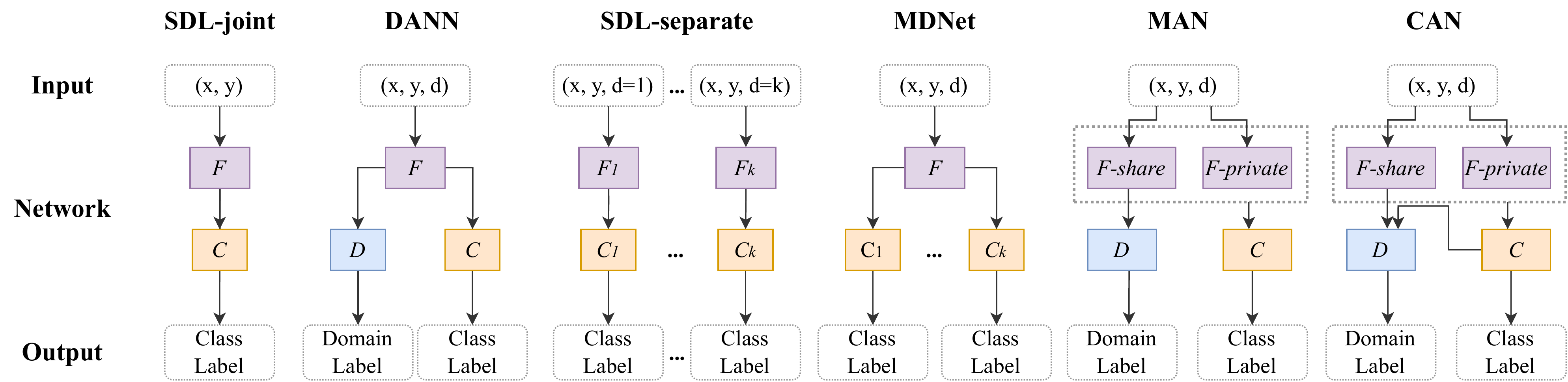}
   \caption{The sketches of different models: $F$ represents feature extractors. $D$ represents domain discriminators. $C$ represents classifiers. $x$, $y$, and $d$ represent the input features, the labels of instances, and the domain ID of the instances, respectively.}
   \label{fig:model_sketches}
\end{figure*}

The neural networks are used as the base model due to their strong ability for representation extraction.
The structures of the models introduced in the original paper are not precisely used due to the difference in the dimensions of inputs.
The macrostructures of the models remain the same as in the original paper.
In other words, the feature extractors (F), classifiers (C), and discriminators (D) are organized in the same way, as sketched in Fig.~\ref{fig:model_sketches}.
Furthermore, the microstructures of different models are the same for the same dataset.
The same module (F, C, or D) of different models have the same depth and number of neurons.
The details can be found in the supplementary materials.

\begin{table}[htbp]
   \caption{The characteristics of the selected strategies. Informativeness (Info.), representativeness (Rep.), batch \& diversity (Div.) and two-stage selection are considered.}
   \centering
   \begin{tabular}{ccccc}
      \toprule
      Strategy    & Info.      & Rep.       & Batch \& Div. & Two-Stage  \\
      \midrule
      Random      & \text{-}   & \text{-}   & \text{-}      & \text{-}   \\
      Uncertainty & \checkmark & \text{-}   & \text{-}      & \text{-}   \\
      EGL         & \checkmark & \text{-}   & \text{-}      & \text{-}   \\
      Coreset     & \text{-}   & \checkmark & \checkmark    & \checkmark \\
      BADGE       & \checkmark & \checkmark & \checkmark    & \checkmark \\
      \bottomrule
   \end{tabular}
   \label{table:strategy-type}
\end{table}

\subsection{Strategies}
\label{sec:compared-strategies}

Most of the strategies in Section~\ref{sec:active-learning-domains} are not appropriate to be utilized in our comparison, even though there are different domains in their settings.
They either heavily depend on models other than neural networks or can only select instances from a single target domain for domain adaptation.
Thus, five conventional single domain AL strategies that can be easily implemented on neural networks are selected for our MDAL pipeline.
The characteristics of the compared strategies are shown in Table~\ref{table:strategy-type}.

\begin{itemize}
   \item
         \textbf{Random}:
         Randomly select instances from each domain.
   \item
         \textbf{Uncertainty}:
         Best vs. Second Best (BvSB) \cite{BvSB}, as an uncertainty measurement, selects instances with the greatest difference in prediction probability between the most and second most likely classes.
   \item
         \textbf{EGL} \cite{EGL,EGL_text}:
         Expected Gradient Length is designed for models that can be optimized by gradients.
         The instances leading to the longest expected gradient length to the last fully connected layer will be selected.
   \item
         \textbf{Coreset} \cite{coreset}:
         Coreset selects instances using a greedy furthest-first search conditioned on the currently labeled examples.
         The distance is calculated by using the output of the penultimate layer from the model.
   \item
         \textbf{BADGE} \cite{BADGE}:
         Batch Active learning by Diverse Gradient Embeddings calculates the gradients of the last fully connected layer.
         A \textit{k}-means++ initialization is applied to the gradients to ensure the representativeness and diversity of the selected batch.
\end{itemize}

\begin{table*}[htbp]
   \caption{The hyperparameters used for the model training and the AL procedure.}
   \centering
   \scalebox{0.95}{
      \begin{tabular}{cccccccccccc}
         \toprule
         \multirow{2}[3]{*}{Datasets} & \multicolumn{7}{c}{Model Training} & \multicolumn{4}{c}{AL Procedure}                                                             \\
         \cmidrule(l){2-8} \cmidrule(l){9-12}
                                      & Optimizer                          & \tabincell{c}{Learning                                                                       \\Rate} & \tabincell{c}{Learning  \\Rate Decay} & \tabincell{c}{Batch \\Size}  & \tabincell{c}{Weight \\decay} & \tabincell{c}{Early \\Stopping}  & \tabincell{c}{Discriminator \\Tradeoff}  & \tabincell{c}{Total \\Budget} & \tabincell{c}{Initial \\Labeled Size} & \tabincell{c}{AL \\Batch Size} & \tabincell{c}{Repeat \\Times}\\
         \midrule
         Amazon                       & Adam                               & 1e-4                             & -     & 128 & 0.05  & 10 & 0.1 & 8000  & 1000 & 1000 & 10 \\
         Office-31                    & Adam                               & 3e-3                             & 0.333 & 128 & 0.001 & 30 & 0.1 & 2400  & 400  & 400  & 10 \\
         Office-Home                  & Adam                               & 1e-4                             & -     & 128 & 0.001 & 10 & 0.1 & 9000  & 1000 & 2000 & 5  \\
         ImageCLEF                    & Adam                               & 3e-3                             & 0.333 & 32  & 0.001 & 25 & 0.1 & 1080  & 180  & 180  & 20 \\
         Digits                       & SGD                                & 1e-2                             & 0.1   & 128 & 0.001 & 15 & 0.1 & 18000 & 2000 & 4000 & 5  \\
         PACs                         & SGD                                & 1e-3                             & 0.1   & 32  & 0.001 & 15 & 0.1 & 8500  & 500  & 2000 & 3  \\
         \bottomrule
      \end{tabular}
   }
   \label{tabel:hyperparameters}
\end{table*}

\subsection{Details of Implementations}
\label{sec:implementation}
All the hyperparameters used for the model training and for the AL procedure are listed in Table \ref{tabel:hyperparameters}.
\subsubsection{Neural Network Training}
Adam and SGD optimizers are used to train the models.
The weight decay term and the cross-entropy loss are used.
A trade-off parameter should be set for balancing the loss of classifiers and discriminators for DANN, MAN, and CAN.
Besides, the early stopping technique is used during the training process.
Most of the hyperparameters are set to be the same for each model on the same dataset to ensure the fairness of the comparison.
\subsubsection{AL Setting}
As a pool-based active learning scenario, the MDAL learning loop continues until the budget is depleted.
A small set of labeled instances (compared to the budget) is randomly selected at the beginning for training an initial model as a warm start.
Relatively large AL batch sizes are set to reduce the overall computation burden, because the smaller the AL batch size, the more times of re-training (more iterations) are performed.
The AL process repeats multiple times to obtain an average performance with a standard deviation.

\subsection{Evaluation Metrics}
\label{sec:metrics}
\subsubsection{Learning curve (micro-average accuracy)}
The micro-average accuracy on the test set is recorded for each model-strategy combination.
As an AL process, the performances are presented as learning curves.
In each curve, the x-axis represents the current cost (the number of labeled instances), and the y-axis represents the accuracy under the current cost.
Each curve represents a model-strategy combination.

\subsubsection{Area under the learning curve (AULC)}
The area under each learning curve can be calculated as an approximate performance evaluation.
AULC is used when we need to compare plenty of combinations simultaneously.
The labeling budget should normalize the area so that the final AULC score is a positive number from 0 to 1.

\section{Results and Analysis}
\label{sec:results}

We conduct the following comparisons to answer the three research questions mentioned in Section~\ref{sec:pipeline}.
First, the models equipped with different information-sharing schemes are compared in Section~\ref{sec:passive}.
Then, Section~\ref{sec:active} compares the AL strategies on each model according to their total performance over domains.
Finally, the per-domain performance is discussed in Section~\ref{sec:robust_domain}.

\subsection{Comparisons over Models}
\label{sec:passive}

We are curious about whether there is a model or an information-sharing scheme naturally suitable for MDAL.
The final performance of MDAL methods is contributed by both the models and the strategies.
A top-performed model under random selection is more likely to perform better than other models under the same AL strategy.
Thus, in this section, the models with different information-sharing schemes are compared with randomly selected labeled training instances.
A model that maintains a good performance in all stages of the whole learning process is desired.

\begin{figure*}[htbp]
   \centering
   \hspace{-0.025\linewidth}
   \subfigure[Office31]{
      \begin{minipage}[b]{0.33\linewidth}
         \includegraphics[width=1\textwidth]{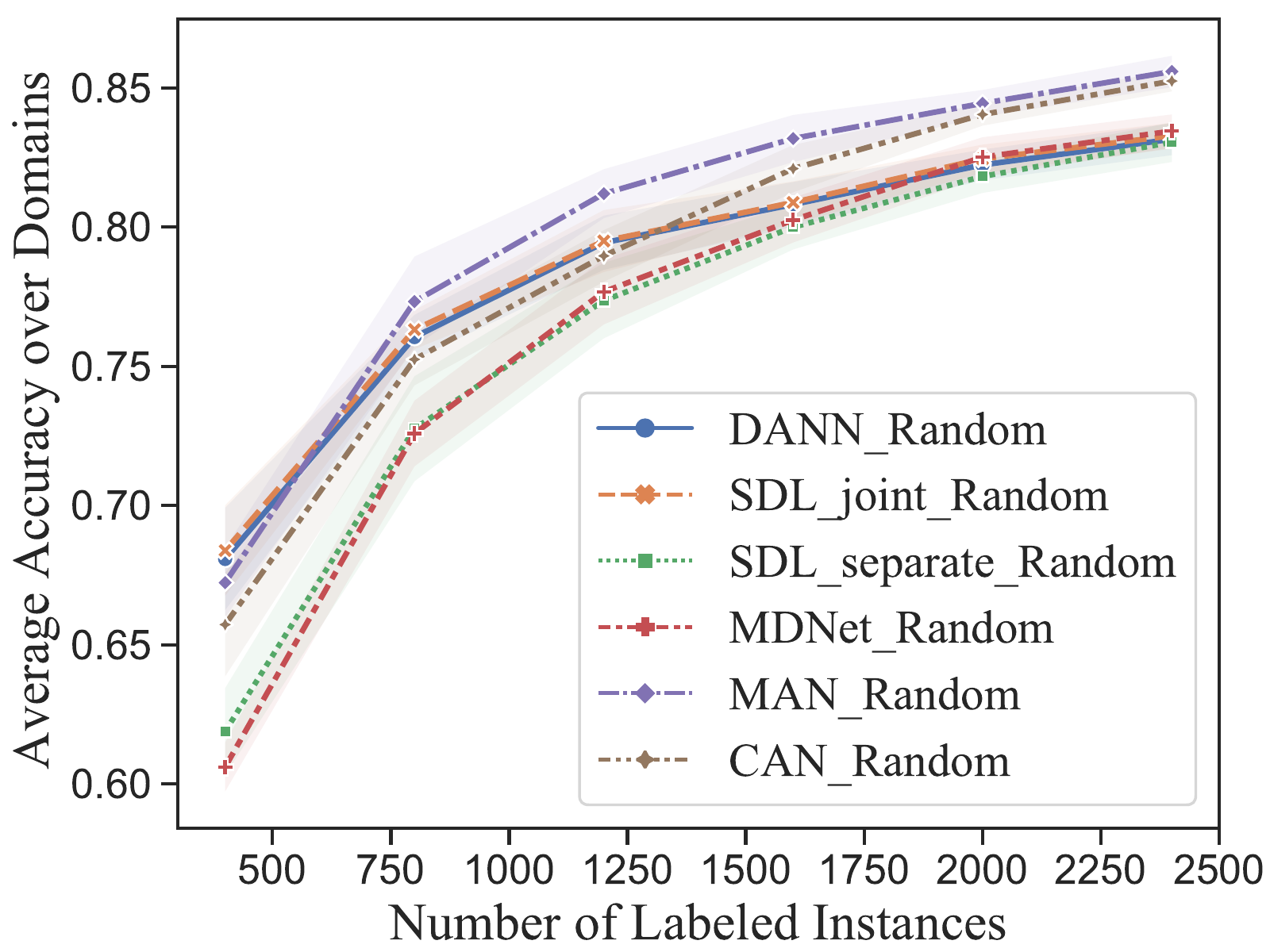}
      \end{minipage}
      \label{fig:shallow-office31}
   }
   \hspace{-0.025\linewidth}
   \subfigure[ImageCLEF]{
      \begin{minipage}[b]{0.33\linewidth}
         \includegraphics[width=1\textwidth]{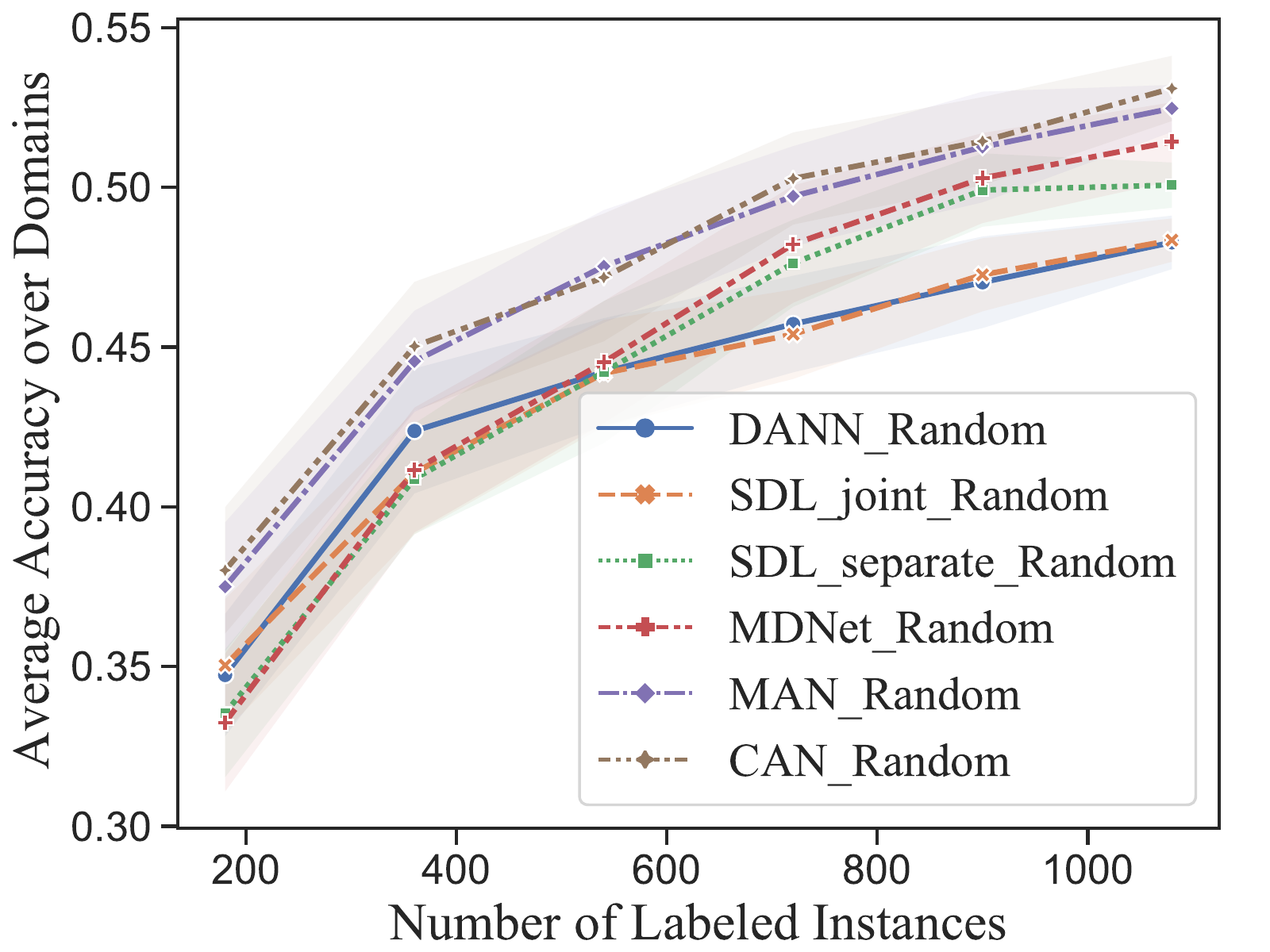}
      \end{minipage}
   }
   \hspace{-0.025\linewidth}
   \subfigure[Amazon]{
      \begin{minipage}[b]{0.33\linewidth}
         \includegraphics[width=1\textwidth]{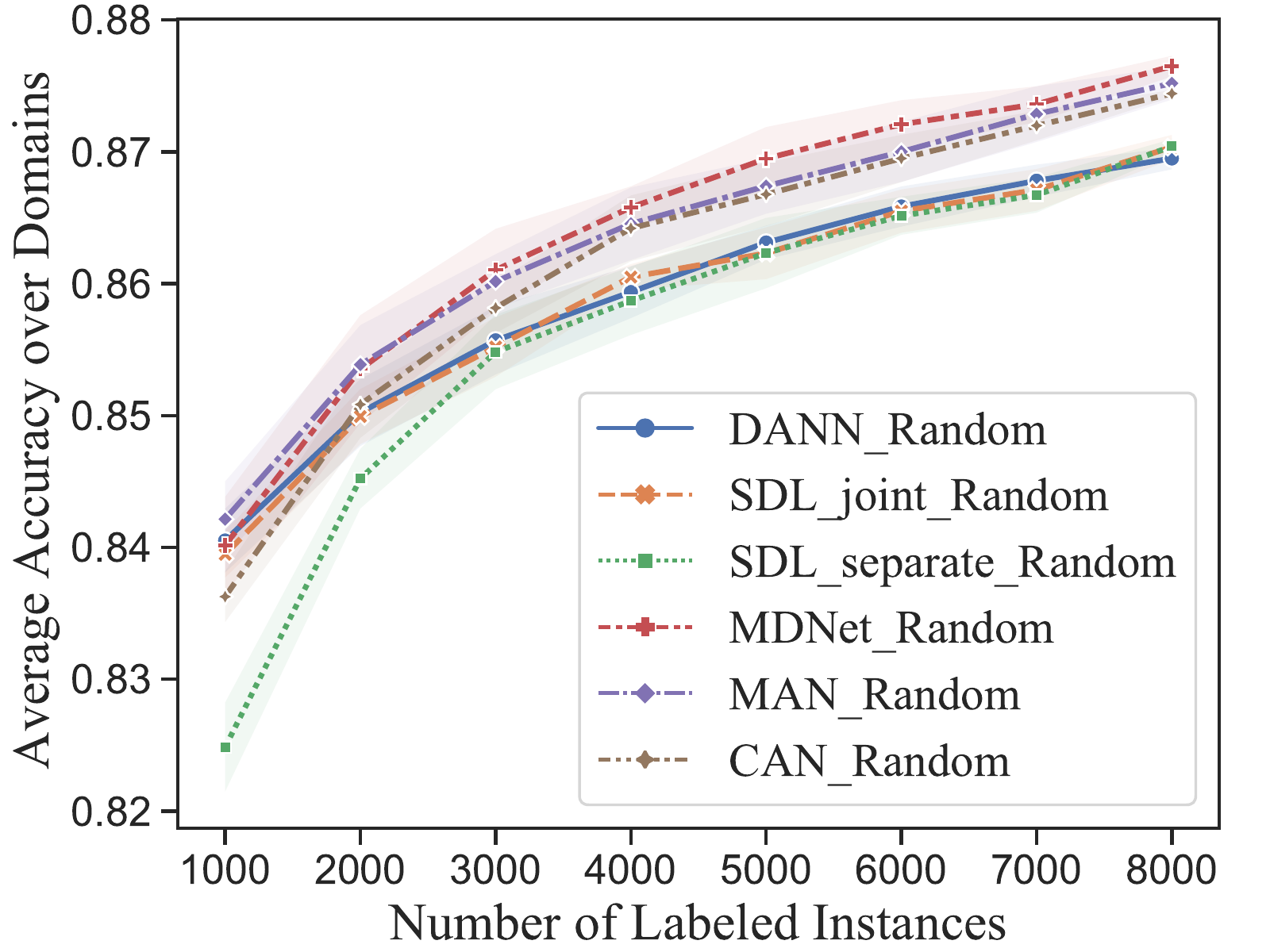}
      \end{minipage}
      \label{fig:shallow-amazon}
   }
   \\
   \hspace{-0.025\linewidth}
   \subfigure[Office-Home]{
      \begin{minipage}[b]{0.33\linewidth}
         \includegraphics[width=1\textwidth]{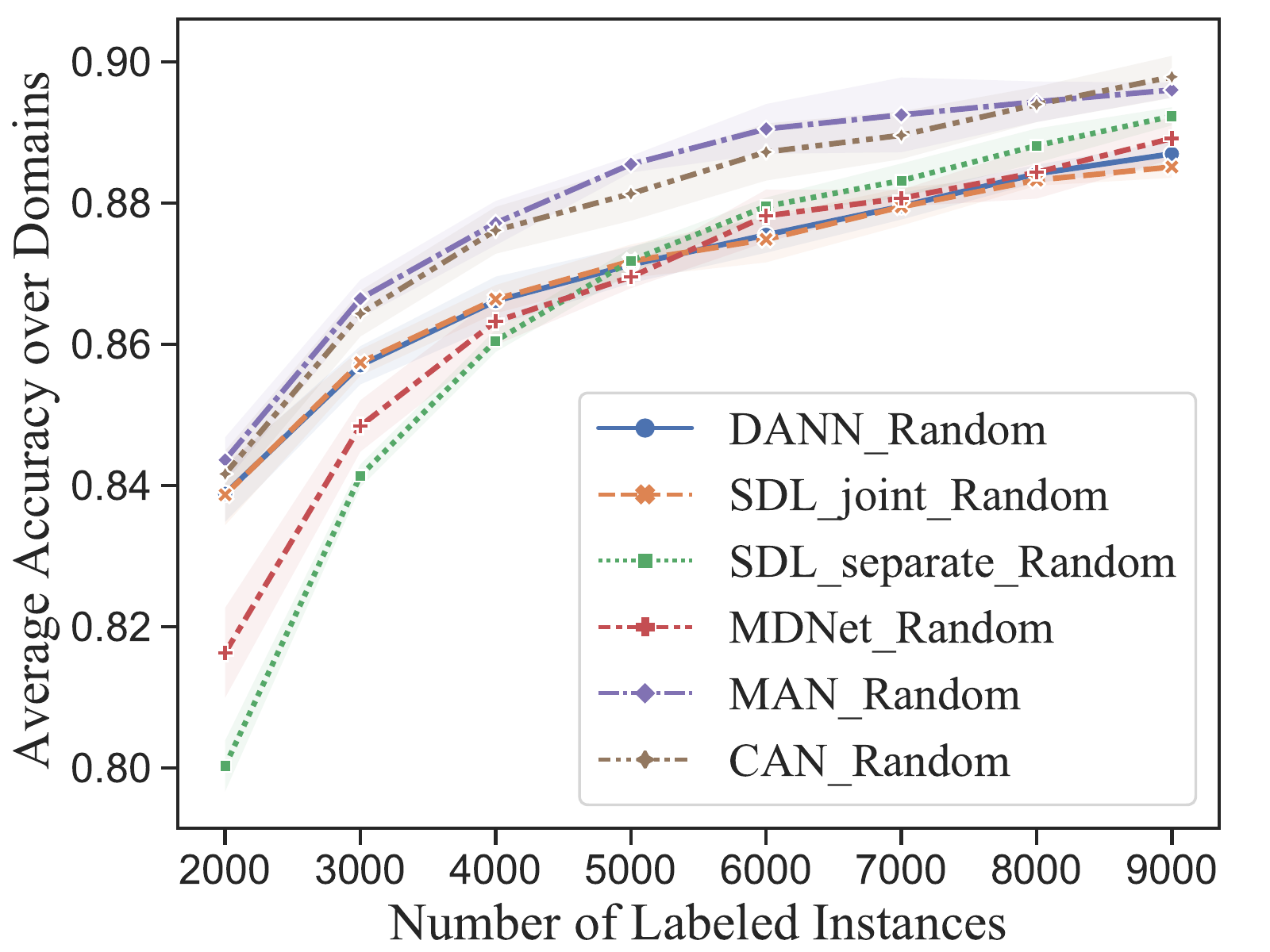}
      \end{minipage}
      \label{fig:shallow-office-home}
   }
   \hspace{-0.025\linewidth}
   \subfigure[Digits]{
      \begin{minipage}[b]{0.33\linewidth}
         \includegraphics[width=1\textwidth]{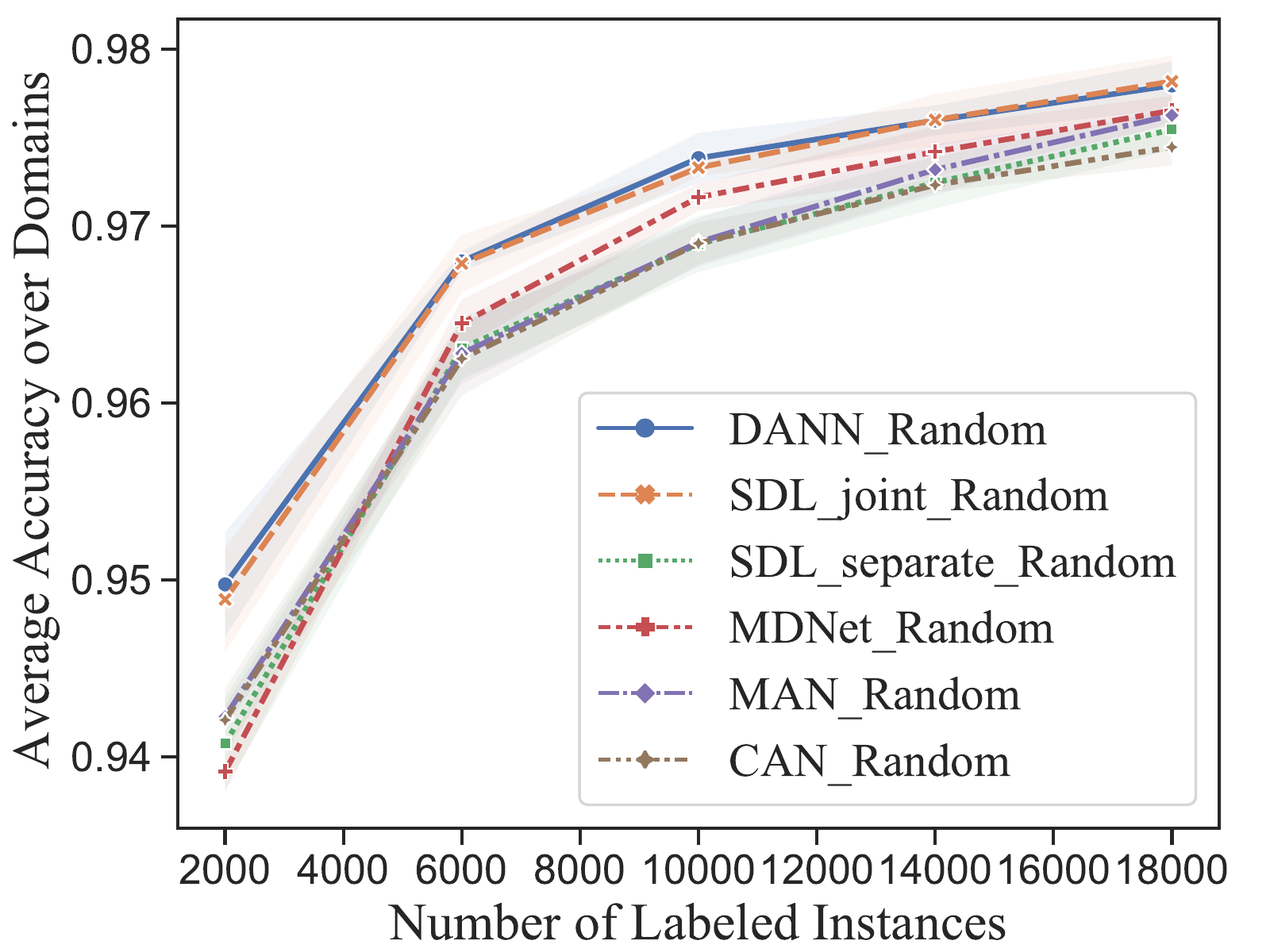}
      \end{minipage}
      \label{fig:deep-digits}
   }
   \hspace{-0.025\linewidth}
   \subfigure[PACs]{
      \begin{minipage}[b]{0.33\linewidth}
         \includegraphics[width=1\textwidth]{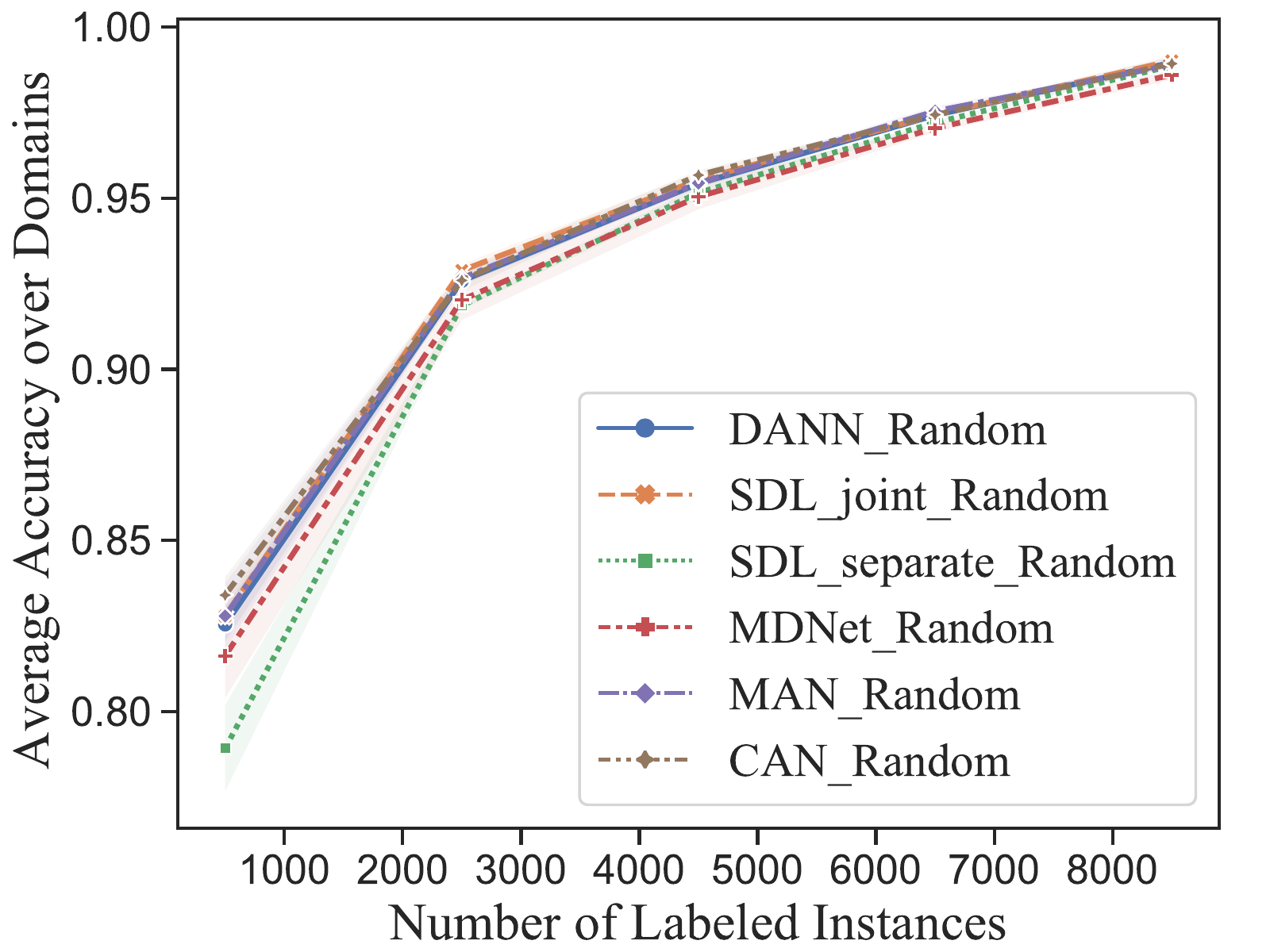}
      \end{minipage}
      \label{fig:deep-pac}
   }
   \caption{The results of different models on six datasets. Models are compared under different number of randomly selected labeled training instances.}
   \label{fig:shallow}
\end{figure*}

Shallow neural networks with a single hidden layer are used for Office31, ImageCLEF, Amazon, and Office-Home datasets.
The features are encoded as vectors in these datasets.
The results on these datasets are plotted in Fig.~\ref{fig:shallow}\subref{fig:shallow-office31}-\subref{fig:shallow-office-home}.
It is easy to find the similar trends from the results on these four datasets.
The whole inference structures of DANN and SDL-joint are trained on labeled instances from all the domains.
Their learning curves have similar trends, and they usually perform well in the beginning but cannot outperform the others in the end.
This is because these two models are relatively well-trained with more labeled instances in the beginning compared to other models.
SDL-separate and MDNet both have domain-specific classifiers.
They usually do not perform well initially, but in the end, they can reach or outperform other models.
In the beginning, the domain-specific classifiers can hardly be well-trained with insufficient labeled instances from the corresponding domains, which brings bad initial performances.
Moreover, MDNet sometimes outperforms SDL-separate too much, as shown in Fig.~\ref{fig:shallow-amazon}.
It comes from the shared extractor and the simpleness of the binary classification task.
MAN and CAN both have share-private inference structures.
Their curves also have similar trends, and their performances are usually good during the whole learning process.
We believe this superiority comes from the share-private structure, which captures the shared information well in the beginning and maintains the domain-private information in the end.
Besides, compared to MAN, CAN sometimes performs clearly worse initially, as shown in Fig.~\ref{fig:shallow}\subref{fig:shallow-office31}~and~\subref{fig:shallow-amazon}.
This is because its conditional distribution matching part cannot be well-trained at the beginning, leading to a negative effect.

Deep neural networks are used for Digits and PACs datasets under raw image features.
The results are plotted in Fig.~\ref{fig:shallow}\subref{fig:deep-digits}~and~\subref{fig:deep-pac}.
We note that on the PACs dataset, the pretrained ResNet18 \cite{ResNet18} is adopted as the feature extractor.
It is easy to find that SDL-joint and DANN perform best and SDL-separate performs worst in both datasets.
MAN and CAN almost perform the worst on the Digits dataset, but they can obtain a top performance on the PACs dataset.
This is because the deep model is less likely well-trained in the beginning without a good initialization (pretrained model) when the labeled instances are very few.

In short, for the first research question, considering the whole learning process, MAN consistently performs well on the shallow networks, and SDL-joint and DANN perform well on the deep networks with the random selected training instances.
With pretrained models, MAN can also obtain the top performance with the deep model.

\begin{table*}[htbp]
   \caption{The area under the learning curve for each model-strategy pair on each dataset. The largest AULC value is \textbf{bolded}.}
   \centering
   \scalebox{1}{
      \begin{tabular}{cccccccc}
         \toprule
         Datasets    & \diagbox{Strategies}{Models} & SDL-joint                & DANN                     & SDL-separate       & MDNet                      & MAN                      & CAN                      \\
         \midrule
                     & Random                       & $ 79.00 \pm 0.44 $       & $78.83 \pm 0.42$         & $76.88 \pm 0.83$   & $ 77.01 \pm 0.57 $         & $ 80.51 \pm 0.48 $       & $ 79.17 \pm 0.41 $       \\
                     & Uncertainty                  & $ 79.98 \pm 0.44 $       & $80.12 \pm 0.41$         & $78.80 \pm 0.49$   & $ 79.03 \pm 0.37 $         & $ 81.94 \pm 0.42 $       & $ 80.50 \pm 0.62 $       \\
         Office-31   & BADGE                        & $ 79.98 \pm 0.46 $       & $80.00 \pm 0.52$         & $78.65 \pm 0.45$   & $ 78.58 \pm 0.34 $         & \pmb{$81.95 \pm 0.37 $}  & $ 80.46 \pm 0.71 $       \\
                     & EGL                          & $ 79.97 \pm 0.60 $       & $79.80 \pm 0.36$         & $78.17 \pm 0.22$   & $ 78.40 \pm 0.48 $         & $ 81.49 \pm 0.24 $       & $ 80.01 \pm 0.58 $       \\
                     & Coreset                      & $ 80.06 \pm 0.44 $       & $79.94 \pm 0.44$         & $79.25 \pm 0.38$   & $ 77.84 \pm 0.50 $         & $ 81.84 \pm 0.31 $       & $ 80.39 \pm 0.47 $       \\
         \midrule
                     & Random                       & $ 85.94 \pm 0.10 $       & $ 85.96 \pm 0.07$        & $ 85.72 \pm 0.13$  & $ 86.48 \pm 0.13   $       & $ 86.39 \pm 0.13     $   & $ 86.24 \pm 0.12     $   \\
                     & Uncertainty                  & $ 86.38 \pm 0.05 $       & $ 86.35 \pm 0.05$        & $ 86.04 \pm 0.07$  & \pmb{$ 86.94 \pm 0.09   $} & $ 86.88 \pm 0.03     $   & $ 86.66 \pm 0.07     $   \\
         Amazon      & BADGE                        & $ 86.34 \pm 0.05 $       & $ 86.34 \pm 0.08$        & $ 85.96 \pm 0.09$  & $ 86.88 \pm 0.07   $       & $ 86.84 \pm 0.07     $   & $ 86.60 \pm 0.07     $   \\
                     & EGL                          & $ 86.37 \pm 0.06 $       & $ 86.38 \pm 0.06$        & $ 86.03 \pm 0.06$  & $ 86.93 \pm 0.10   $       & $ 86.89 \pm 0.06     $   & $ 86.69 \pm 0.05     $   \\
                     & Coreset                      & $ 86.26 \pm 0.08 $       & $ 86.26 \pm 0.05$        & $ 85.95 \pm 0.08$  & $ 86.83 \pm 0.09   $       & $ 86.74 \pm 0.05     $   & $ 86.41 \pm 0.06     $   \\
         \midrule
                     & Random                       & $ 87.07 \pm 0.12 $       & $ 87.09 \pm 0.10 $       & $ 86.72 \pm 0.05 $ & $ 86.81 \pm 0.13 $         & $ 88.23 \pm 0.20 $       & $ 88.03 \pm 0.18 $       \\
                     & Uncertainty                  & $ 87.78 \pm 0.12 $       & $ 87.81 \pm 0.11 $       & $ 87.78 \pm 0.09 $ & $ 87.82 \pm 0.03 $         & $ 88.90 \pm 0.06 $       & $ 88.78 \pm 0.11 $       \\
         Office-Home & BADGE                        & $ 87.77 \pm 0.05 $       & $ 87.75 \pm 0.12 $       & $ 87.75 \pm 0.03 $ & $ 87.72 \pm 0.06 $         & \pmb{$ 88.99 \pm 0.07 $} & $ 88.75 \pm 0.09 $       \\
                     & EGL                          & $ 87.85 \pm 0.14 $       & $ 87.79 \pm 0.08 $       & $ 87.72 \pm 0.02 $ & $ 87.76 \pm 0.07 $         & $ 88.84 \pm 0.08 $       & $ 88.77 \pm 0.04 $       \\
                     & Coreset                      & $ 87.75 \pm 0.13 $       & $ 87.73 \pm 0.09 $       & $ 87.86 \pm 0.05 $ & $ 87.58 \pm 0.08 $         & $ 88.97 \pm 0.13 $       & $ 88.68 \pm 0.08 $       \\
         \midrule
                     & Random                       & $ 43.92 \pm 0.69 $       & $ 44.17 \pm 0.84 $       & $ 44.89 \pm 0.93 $ & $ 45.31 \pm 1.07 $         & $ 47.61 \pm 0.85 $       & $ 47.90 \pm 0.84 $       \\
                     & Uncertainty                  & $ 43.55 \pm 0.67 $       & $ 43.99 \pm 0.80 $       & $ 44.89 \pm 0.83 $ & $ 45.52 \pm 0.87 $         & $ 47.49 \pm 0.83 $       & $ 47.96 \pm 1.09 $       \\
         ImageCLEF   & BADGE                        & $ 43.80 \pm 0.95 $       & $ 44.14 \pm 0.76 $       & $ 45.20 \pm 0.98 $ & $ 45.42 \pm 0.85 $         & $ 47.93 \pm 0.66 $       & \pmb{$ 48.47 \pm 1.13 $} \\
                     & EGL                          & $ 43.15 \pm 0.89 $       & $ 43.35 \pm 0.95 $       & $ 44.13 \pm 1.11 $ & $ 44.51 \pm 0.79 $         & $ 47.03 \pm 0.89 $       & $ 47.17 \pm 1.21 $       \\
                     & Coreset                      & $ 43.27 \pm 1.01 $       & $ 43.38 \pm 0.72 $       & $ 44.68 \pm 0.82 $ & $ 44.32 \pm 0.93 $         & $ 47.49 \pm 0.76 $       & $ 48.12 \pm 1.05 $       \\
         \midrule
                     & Random                       & $ 97.02 \pm 0.10 $       & $ 97.04 \pm 0.06 $       & $ 96.57 \pm 0.09 $ & $ 96.70 \pm 0.06 $         & $ 96.61 \pm 0.07 $       & $ 96.55 \pm 0.08 $       \\
                     & Uncertainty                  & $ 97.93 \pm 0.08 $       & $ 97.97 \pm 0.07 $       & $ 97.67 \pm 0.05 $ & $ 97.53 \pm 0.06 $         & $ 97.75 \pm 0.03 $       & $ 97.67 \pm 0.05 $       \\
         Digits      & BADGE                        & $ 97.97 \pm 0.04 $       & $ 97.99 \pm 0.06 $       & $ 97.71 \pm 0.04 $ & $ 97.60 \pm 0.04 $         & $ 97.78 \pm 0.04 $       & $ 97.67 \pm 0.06 $       \\
                     & EGL                          & $ 97.96 \pm 0.08 $       & \pmb{$ 98.00 \pm 0.04 $} & $ 97.67 \pm 0.06 $ & $ 97.60 \pm 0.03 $         & $ 97.74 \pm 0.02 $       & $ 97.64 \pm 0.04 $       \\
                     & Coreset                      & $ 97.73 \pm 0.05 $       & $ 97.79 \pm 0.06 $       & $ 97.44 \pm 0.04 $ & $ 97.40 \pm 0.03 $         & $ 97.52 \pm 0.02 $       & $ 97.41 \pm 0.04 $       \\
         \midrule
                     & Random                       & $ 94.19 \pm 0.07 $       & $ 94.05 \pm 0.08 $       & $ 93.28 \pm 0.13 $ & $ 93.56 \pm 0.28 $         & $ 94.12 \pm 0.04 $       & $ 94.22 \pm 0.16 $       \\
                     & Uncertainty                  & \pmb{$ 96.34 \pm 0.03 $} & $ 96.32 \pm 0.07 $       & $ 95.31 \pm 0.16 $ & $ 95.94 \pm 0.21 $         & $ 96.09 \pm 0.17 $       & $ 96.28 \pm 0.07 $       \\
         PACs        & BADGE                        & $ 96.26 \pm 0.03 $       & $ 96.18 \pm 0.08 $       & $ 95.21 \pm 0.19 $ & $ 95.96 \pm 0.20 $         & $ 95.96 \pm 0.17 $       & $ 96.00 \pm 0.11 $       \\
                     & EGL                          & $ 96.31 \pm 0.08 $       & $ 96.22 \pm 0.02 $       & $ 95.08 \pm 0.18 $ & $ 95.96 \pm 0.20 $         & $ 95.89 \pm 0.04 $       & $ 95.96 \pm 0.30 $       \\
                     & Coreset                      & $ 94.63 \pm 0.25 $       & $ 93.21 \pm 0.17 $       & $ 93.87 \pm 0.20 $ & $ 93.60 \pm 0.30 $         & $ 94.26 \pm 0.41 $       & $ 94.01 \pm 1.05 $       \\
         \bottomrule
      \end{tabular}
   }
   \label{tabel:all-performance}
\end{table*}

\subsection{Comparisons over Strategies}
\label{sec:active}

All model-strategy combinations are compared on all six datasets in this section.
We are curious about whether AL can bring improvements over the random selection and whether there is any strategy that significantly outperforms the others.
The performances on six datasets are presented in Table~\ref{tabel:all-performance} in terms of AULC.
The raw learning curves can be found in the supplementary materials.
From the results, AL brings clear improvements to almost all the datasets.
However, it is hard to pick one of the strategies that significantly outperforms the others.

Uncertainty, as the most naive AL strategy, consistently obtains good performances in terms of AULC and the learning curves.
On most datasets and models, Uncertainty performs competitively with other state-of-the-art strategies.
The existence of multiple domains might diversify the selection and improve the performance of Uncertainty.
Besides, as a score-based strategy, Uncertainty evaluates the instances solely by the outputs of the models and needs no additional gradients or representations, making it the quickest strategy.
We can claim that Uncertainty is the best strategy for the MDAL pipeline, considering the performance and the evaluation time.

In addition, the final performance of MDAL methods is contributed by both the models and the strategies.
Considering the models, Uncertainty with MAN can achieve top performance on almost all the datasets.
The combination of the well-performed model and strategy is reasonably more likely to perform better than others.
Even though MAN performs poorly with the random selection on the Digits dataset, with Uncertainty, MAN can still obtain a medium performance, as shown in Fig.~\ref{fig:digits-overall}.

In short, no strategy significantly outperforms the others.
However, the naive computationally efficient Uncertainty strategy can perform competitively with other state-of-the-art strategies.
Especially with the MAN model, Uncertainty can consistently obtain top performance in most cases.

\subsection{Comparisons over Domains}
\label{sec:robust_domain}

In this section, we try to find out whether the model-strategy combinations that obtain the excellent overall performance in the previous section can consistently perform well in each domain.
Although the overall performance is primarily concerned in MDAL, a model-strategy combination that performs significantly worse on several domains is not acceptable.
Specifically, the performance of the Uncertainty strategy is discussed, especially with the MAN and the SDL-joint models for their good performance in Section~\ref{sec:compared-models}.

\begin{figure*}[htbp]
   \centering
   \hspace{-0.035\linewidth}
   \subfigure[Digits]{
      \begin{minipage}[b]{0.34\linewidth}
         \includegraphics[width=1\textwidth]{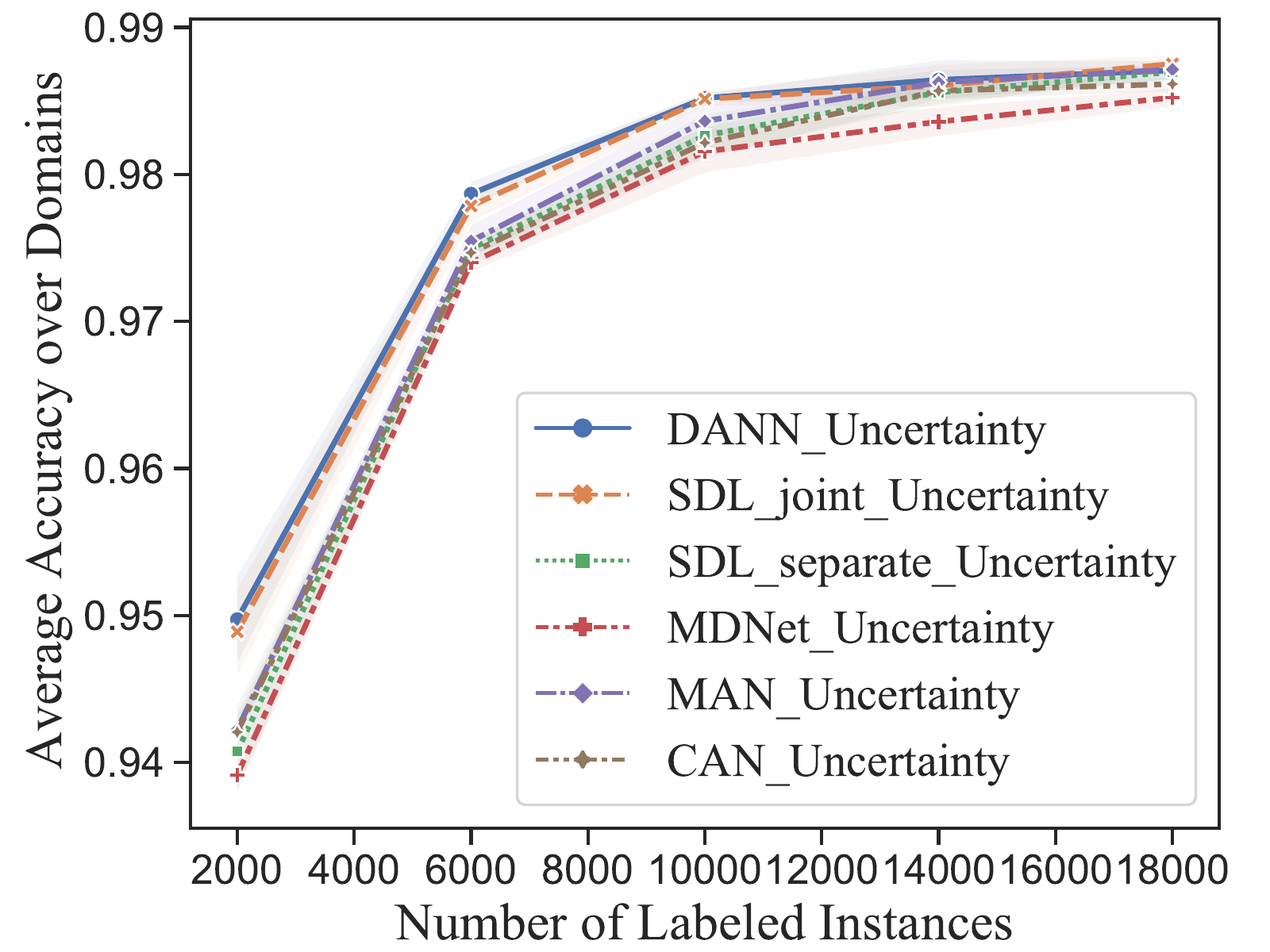}
      \end{minipage}
      \label{fig:digits-overall}
   }
   \hspace{-0.035\linewidth}
   \subfigure[MNIST domain]{
      \begin{minipage}[b]{0.34\linewidth}
         \includegraphics[width=1\textwidth]{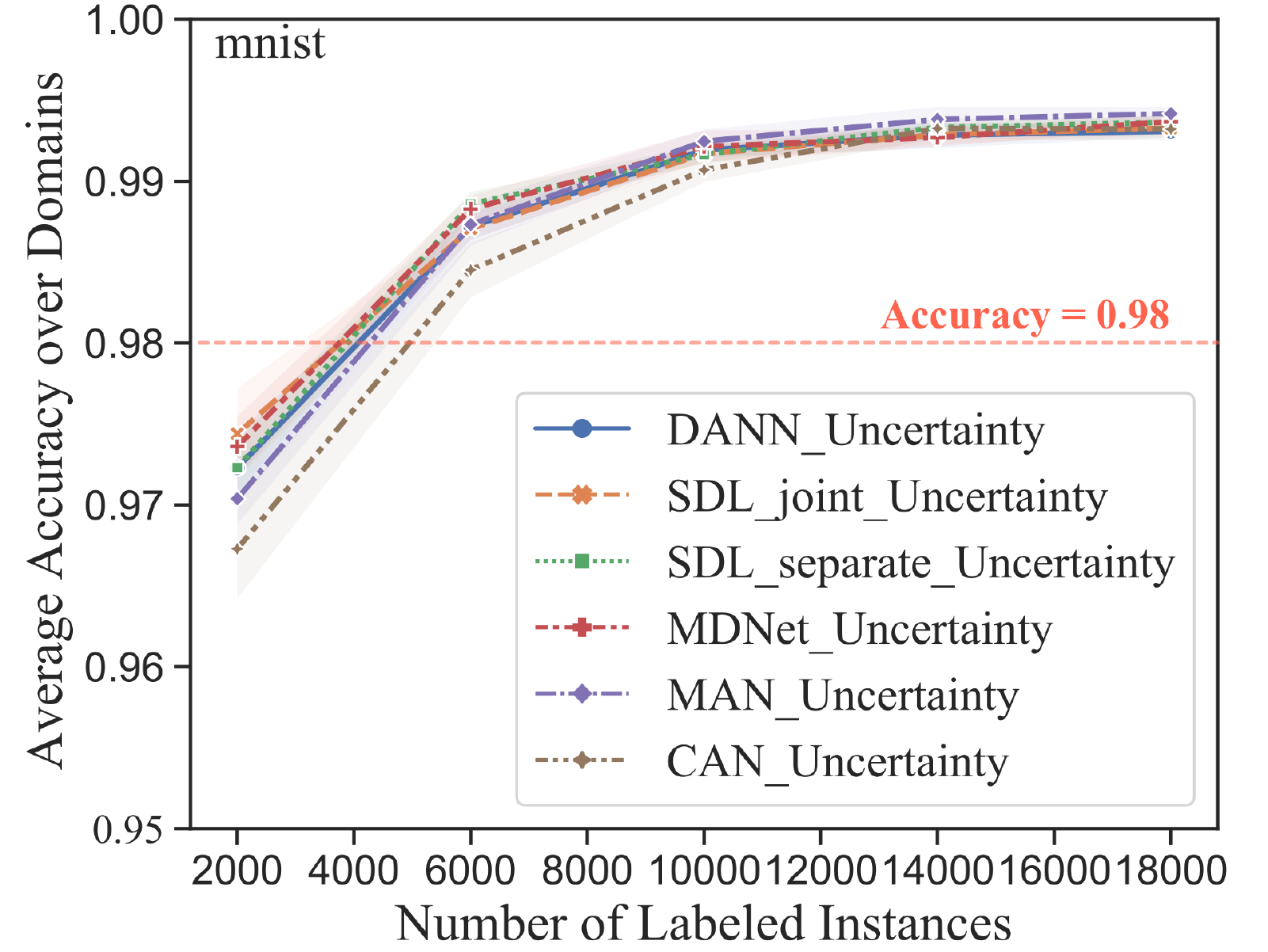}
      \end{minipage}
      \label{fig:digits-mnist}
   }
   \hspace{-0.035\linewidth}
   \subfigure[MNIST-M domain]{
      \begin{minipage}[b]{0.34\linewidth}
         \includegraphics[width=1\textwidth]{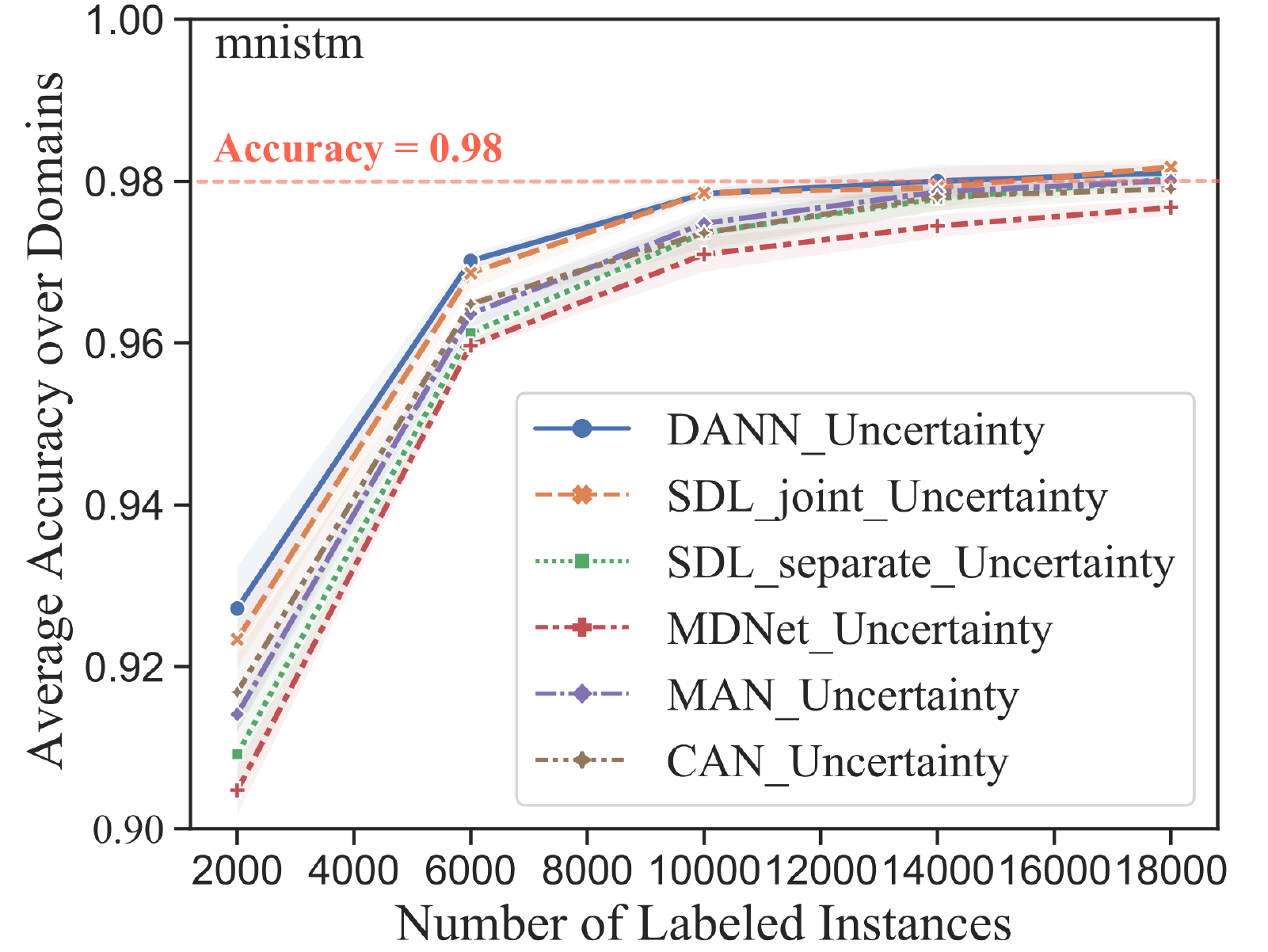}
      \end{minipage}
      \label{fig:digits-mnistm}
   }
   \caption{The results in each domain from the Digits dataset.}
   \label{fig:digit-result-domain}
\end{figure*}

On the Digits dataset, the results of the Uncertainty strategy on both domains are plotted in Fig.~\ref{fig:digit-result-domain}.
The classification task on MNIST-M is supposed to be harder due to the more complex image backgrounds.
The accuracy on MNIST-M is relatively lower than MNIST, as we expected.
On MNIST-M, all the combinations obtain a near 98\% accuracy when the budgets are depleted.
However, on MNIST, the accuracy of 98\% has been exceeded with less than 6000 overall labeled instances.
Besides, the model-strategy combinations perform more concentrated on MNIST than on MNIST-M.
These facts reflect the difficulty of the task on MNIST-M.
SDL-joint with Uncertainty, which obtains the top overall performance on this dataset, achieves an above-average performance on the easier MNIST domain and outperforms the others on the harder MNIST-M domain.
Besides, MAN with Uncertainty performs worse on this dataset compared to its performance on the rest five datasets.
It can still obtain good performance on MNIST and medium performance on MNIST-M.

The performances of different domains on the rest datasets can be found in the supplementary materials.
On these datasets, MAN with Uncertainty usually performs well in most domains, and it can at least obtain above-average performance in the worst performed domain.

In conclusion, the model-strategy combinations (such as MAN-Uncertainty \& SDL-joint-Uncertainty) with the best overall performance usually cannot consistently perform best on all the domains.
However, they can at least obtain medium performance in the worst performed domain in the worse case.
Hence, such combinations can be applied to real scenarios since domain performance can be ensured.

\section{Deeper Investigations}
\label{sec:investigations}

This section analyzes MAN and SDL-joint models and the Uncertainty strategy, which obtain superior performances in the previous comparison.

\begin{figure}[tbp]
   \centering
   \hspace{-0.05\linewidth}
   \subfigure[MAN analysis]{
      \begin{minipage}[b]{0.68\linewidth}
         \includegraphics[width=1\textwidth]{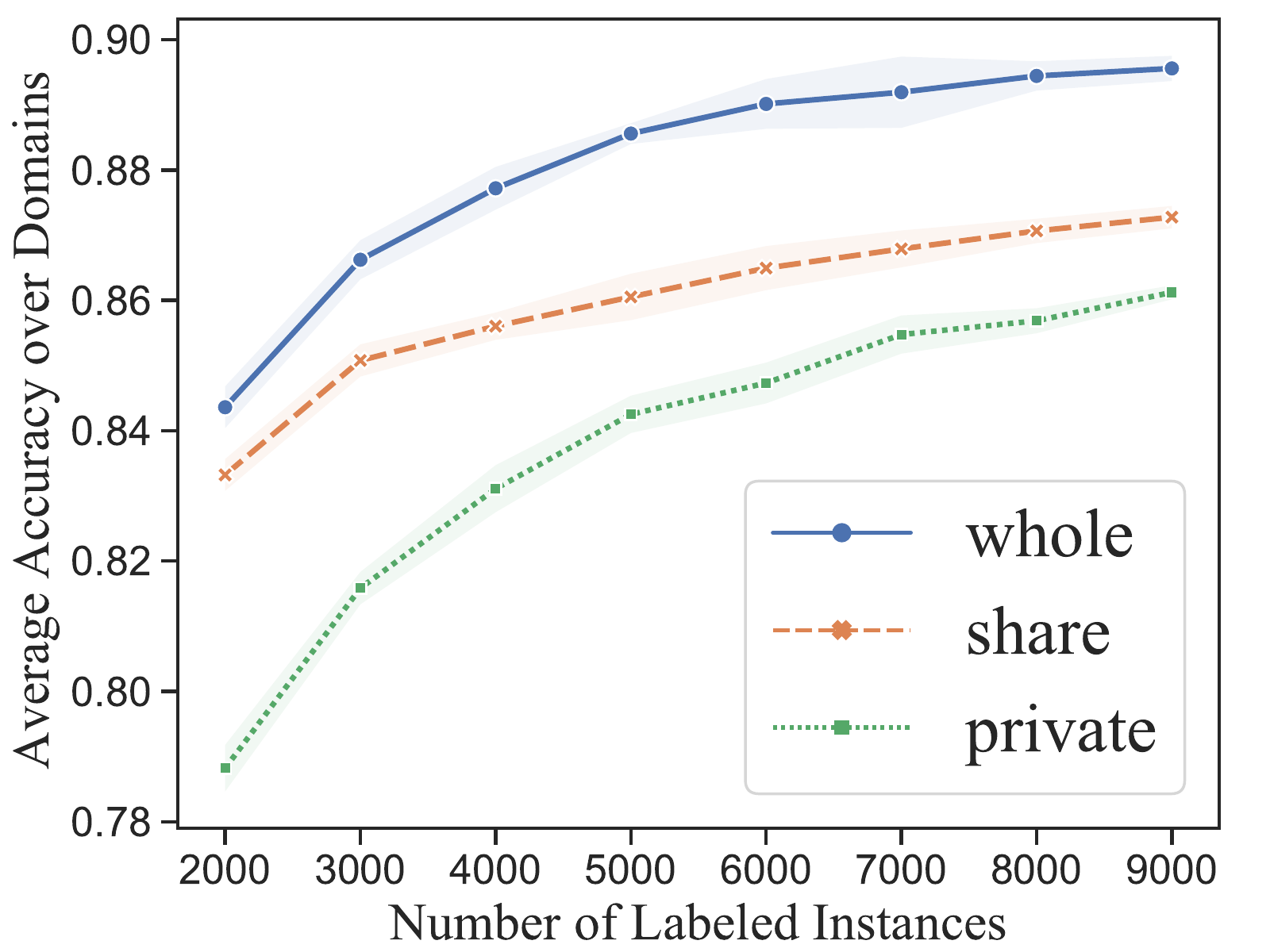}
      \end{minipage}
      \label{fig:man-analysis}
   }
   \hspace{-0.08\linewidth}
   \subfigure[Uncertainty analysis]{
      \begin{minipage}[b]{0.34\linewidth}
         \includegraphics[width=1\textwidth]{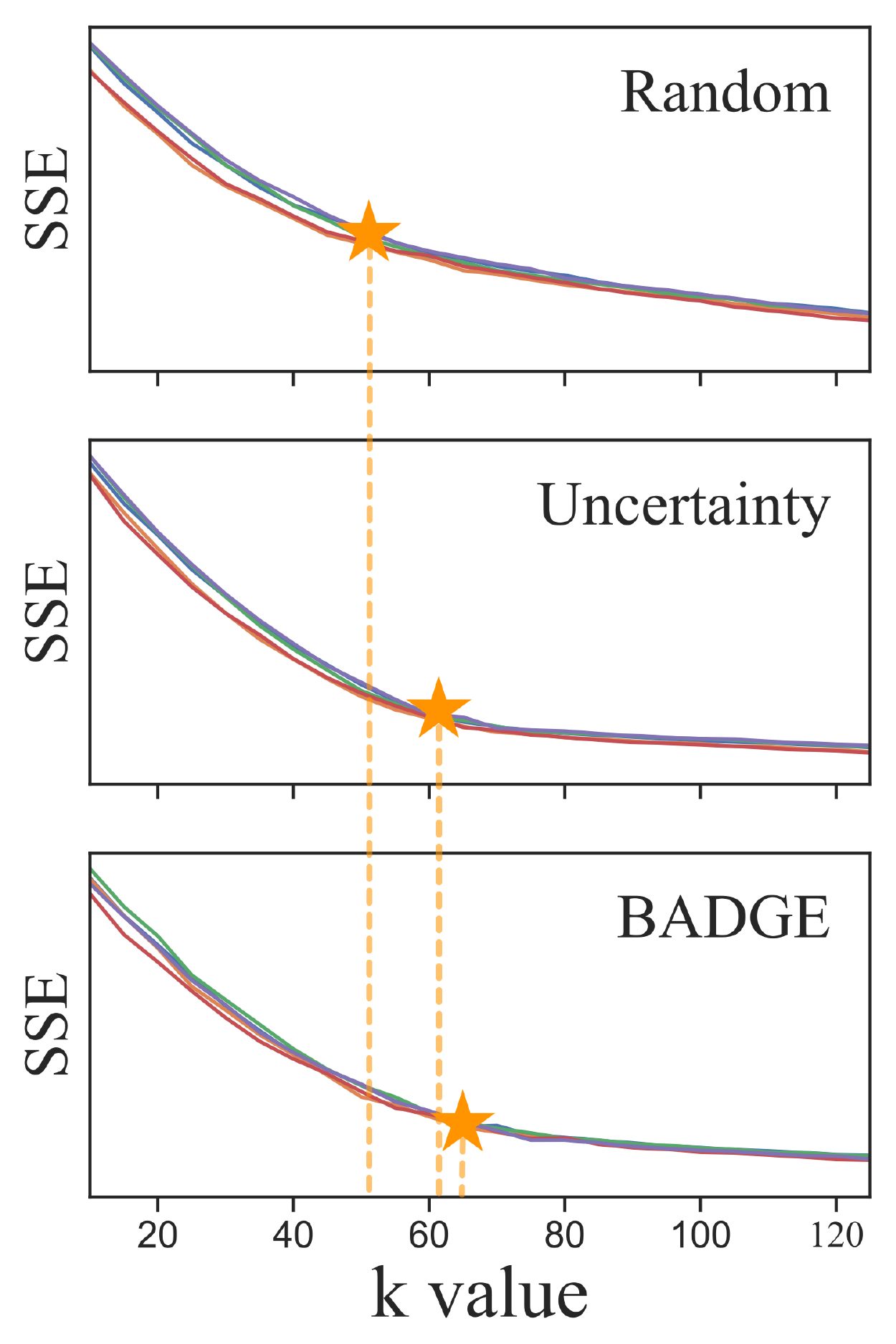}
      \end{minipage}
      \label{fig:uncertainty-analysis}
   }
   \caption{The analysis of MAN and Uncertainty for their superior performance.}
   \label{fig:analysis}
\end{figure}

\textbf{The share-private structure guarantees the good overall performance of MAN.}
The shared feature extractor ensures good initial performance, and the domain-specific extractor captures the domain-specific information at the latter stage of the learning.
An ablation study is made on the Office-Home dataset to verify the utilities of the share-private structure.
The trained MAN model is separated into a shared part and a private part, and each part can make predictions independently.
The test performances of both parts are shown in Fig.~\ref{fig:man-analysis}.
At the beginning, the shared part dominates the performance, and the performance discrepancy between the shared part and the whole model is slight.
The discrepancy increases with more labeled training instances.
The growth rate of the private part is higher than the shared part, and the performances of both parts get closer at the end of the AL process.
The overall performance at the latter stage relies more on the private part than at the former stage since solely using the shared part cannot get comparable performance with the whole model.

\textbf{Without pretrained models, joint training performs well in deep models on raw image datasets.}
SDL-joint and DANN obtain the best performances with deep networks on the Digits and the PACs datasets.
When the model is deep, the information from different domains can be captured by one single inference structure as long as there are no conflicts among domains.
The single inference structure with relatively more labeled training instances (instead of separating training data for each domain) can easily learn a better feature extractor and a better classifier.
Thus, SDL-joint outperforms other MDL models on the Digits dataset.
However, when a good model initialization (pretrained feature extractor) is used in the PACs dataset, SDL-joint's performance improvement relative to MDL models disappears as shown in Fig.~\ref{fig:deep-pac}.

Besides, SDL-joint cannot handle domain conflicts, where similar items on different domains might have different ground truth labels.
For instance, “It is hot” is a positive review for a heater rather than a refrigerator in a sentiment classification task.
In this case, SDL-joint is impossible to obtain correct predictions for both domains because the outputs will remain the same.
However, domain conflicts are not common in image classification, which avoids the performance degradation in this comparison.
In some multi-domain natural language processing tasks, deep MDL models consistently obtain better performances than SDL-joint \cite{MAN, CAN}.

\textbf{The weakness of the Uncertainty strategy in redundant selection is mitigated in MDAL.}
The naive Uncertainty strategy surprisingly performs well in most cases, while it usually performs worse than other state-of-the-art strategies in the conventional single domain AL.
The biggest weakness of Uncertainty is selecting similar instances in a single batch \cite{BADGE}.
We believe this weakness of Uncertainty is relieved by MDAL, where the existence of domains brings higher intra-batch diversity in uncertainty selections.
The diversity of the selected instances by Random, Uncertainty, and BADGE strategy is analyzed.
Considering the first selected batch, the inducted gradients for the last fully connected layer parameters are taken as the embeddings for the instances.
In terms of the interpretability of neural networks \cite{Interpretability}, these gradients reveal the influences from different training instances to the model.
The appropriate number of clusters $k$ in $k$-means can be used to evaluate the diversity of the batch selection.
The $k$ is decided by the elbow method, which takes the $k$ at the turning point of the loss-$k$ curve, as shown in Fig.~\ref{fig:uncertainty-analysis}.
BADGE is supposed to have the highest diversity on the gradient embedding.
The $k$ value of the Uncertainty selection is close to BADGE, which supports the diversity of the Uncertainty selection in MDAL.

\section{Conclusion}
\label{sec:conclusion}

In this work, a comparative study is made for MDAL.
At first, we provide a formal definition of MDAL.
An exhaustive review of the relevant fields is provided subsequently.
Furthermore, we construct an MDAL pipeline as an off-the-shelf solution.
In the pipeline, the combinations of six models from the MDL setting with five AL strategies are compared on six datasets.

The comparison of models reveals the superiority of MAN with the shallow neural networks and the superiority of SDL-joint with the deep neural networks.
Then, Uncertainty performs competitively to the state-of-the-art strategies in the comparison of AL strategies.
Uncertainty can consistently obtain top performance in most datasets with the MAN model.
The combinations obtaining good overall performances can also be robust in different domains, where their in-domain performances are at least above-average.
In the end, their superiority is further investigated.
We recommend using MAN with Uncertainty in applications of MDAL due to the good performance in the experiments.

In the future, this work can be extended in the following directions:
(1) The evaluations of the conventional AL strategies in the pipeline are solely based on the outputs of MDL models. They are not optimal for the whole learning system since the instances most beneficial to the current model (for the current domain) might not bring the most improvement to the overall performance. A new ad-hoc MDAL strategy that can explicitly evaluate the domain-shared and domain-private informativeness should be developed.
(2) The current pipeline has not set importance weights to specific domains, while the importance of distinct domains can be different in real life. It is interesting to update the current pipeline to handle the weights during the training and querying.
(3) Considering the newly added domains in real applications, it is interesting to generalize this pipeline to an active domain generalization problem or a multi-source active domain adaptation problem, where the performance on the new domains is concerned.
(4) The relations among domains are only revealed through the MDL models. It is interesting to include other correlation measurements in the pipeline.


\bibliographystyle{IEEEtran}
\bibliography{mdal}


\vspace{11pt}

\begin{IEEEbiography}[{\includegraphics[width=1in,height=1.25in,clip,keepaspectratio]{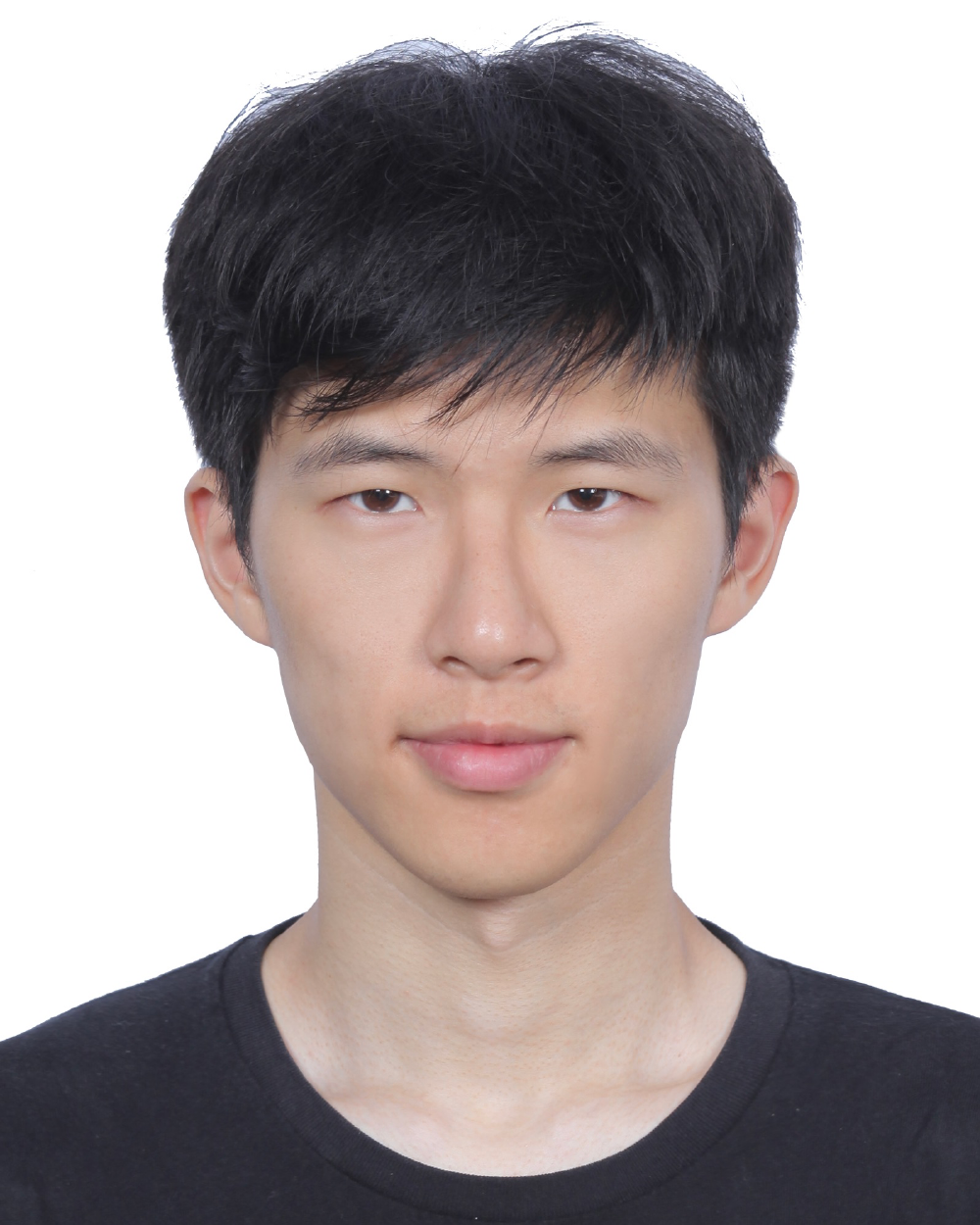}}]{Rui He}
received the B.Sc. degree from the Department of Physics, Southern University of Science and Technology, Shenzhen, China, in 2018. 
He is currently pursuing the Ph.D. degree in the Department of Computer Science and Engineering, Southern University of Science and Technology, jointly with the School of Computer Science, University of Birmingham, Edgbaston, Birmingham, UK. 
His current research interests are active machine learning and multi-domain learning.
\end{IEEEbiography}

\vspace{11pt}

\begin{IEEEbiography}[{\includegraphics[width=1in,height=1.25in,clip,keepaspectratio]{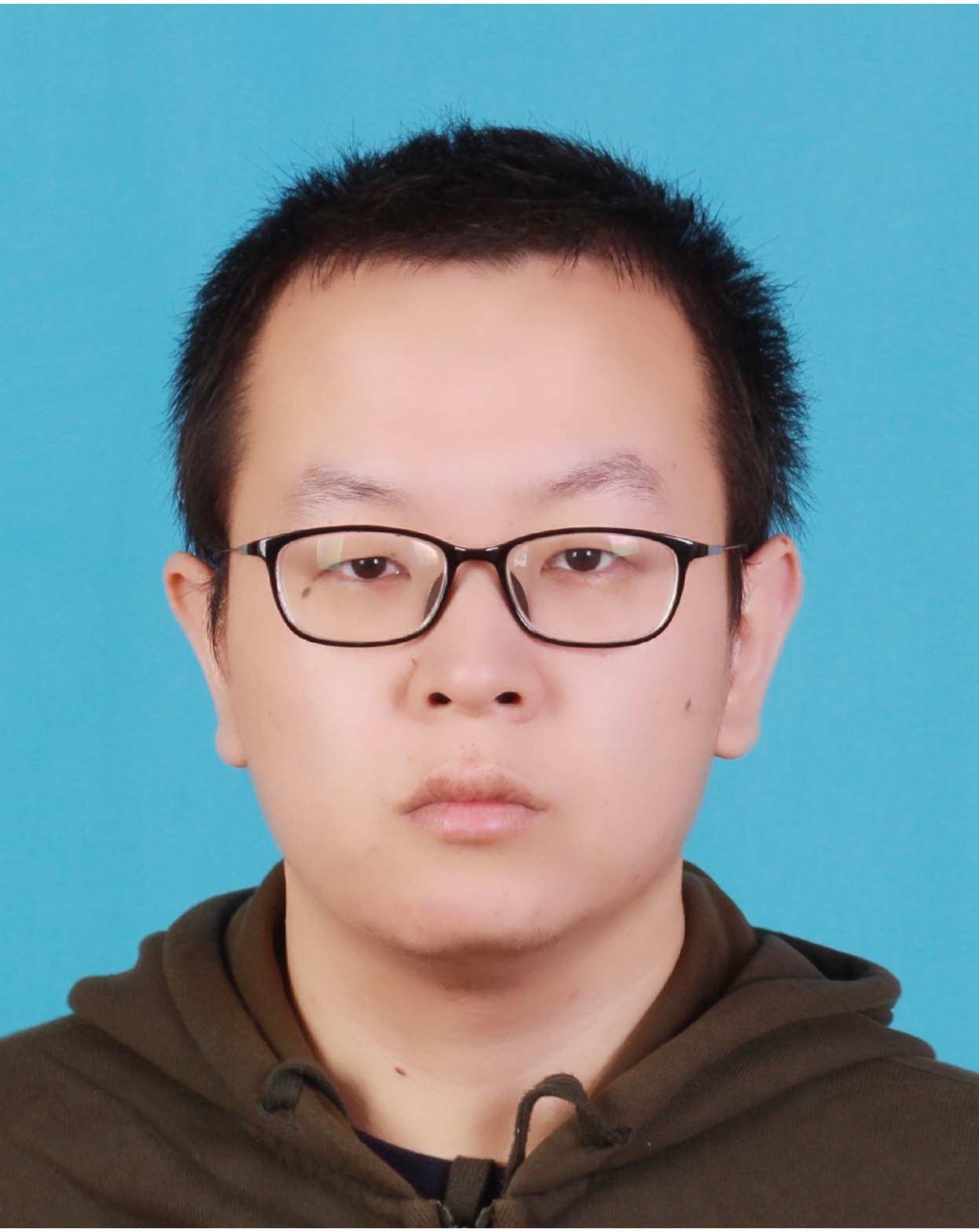}}]{Shengcai Liu}
(M’20) received the B.Eng. degree and Ph.D. degree in computer science and technology from the University of Science and Technology of China (USTC), Hefei, China, in 2014 and 2020, respectively. From 2021, He has been working as a Research Assistant Professor at the Department of Computer Science and Engineering, Southern University of Science and Technology, Shenzhen, China. His research interests include automatic algorithm evolution, combinatorial optimization, and their applications in Computer Vision, Natural Language Processing. He has published more than 10 papers in top-tier refereed international conferences and journals.
\end{IEEEbiography}

\vspace{11pt}

\begin{IEEEbiography}[{\includegraphics[width=1in,height=1.25in,clip,keepaspectratio]{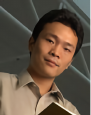}}]{Shan He}
is a Senior Lecturer (Tenured Associate Professor) in  School of Computer Science, the University of Birmingham. He is also an affiliate of the Centre for Computational Biology. His research interests include complex networks, machine learning, optimisation and their applications to biomedicine and drug discovery. Shan is an Associate Editor of IEEE Transactions on Nanobioscience and Complex \& Intelligent Systems (Springer).  
\end{IEEEbiography}

\vspace{11pt}

\begin{IEEEbiography}[{\includegraphics[width=1in,height=1.25in,keepaspectratio]{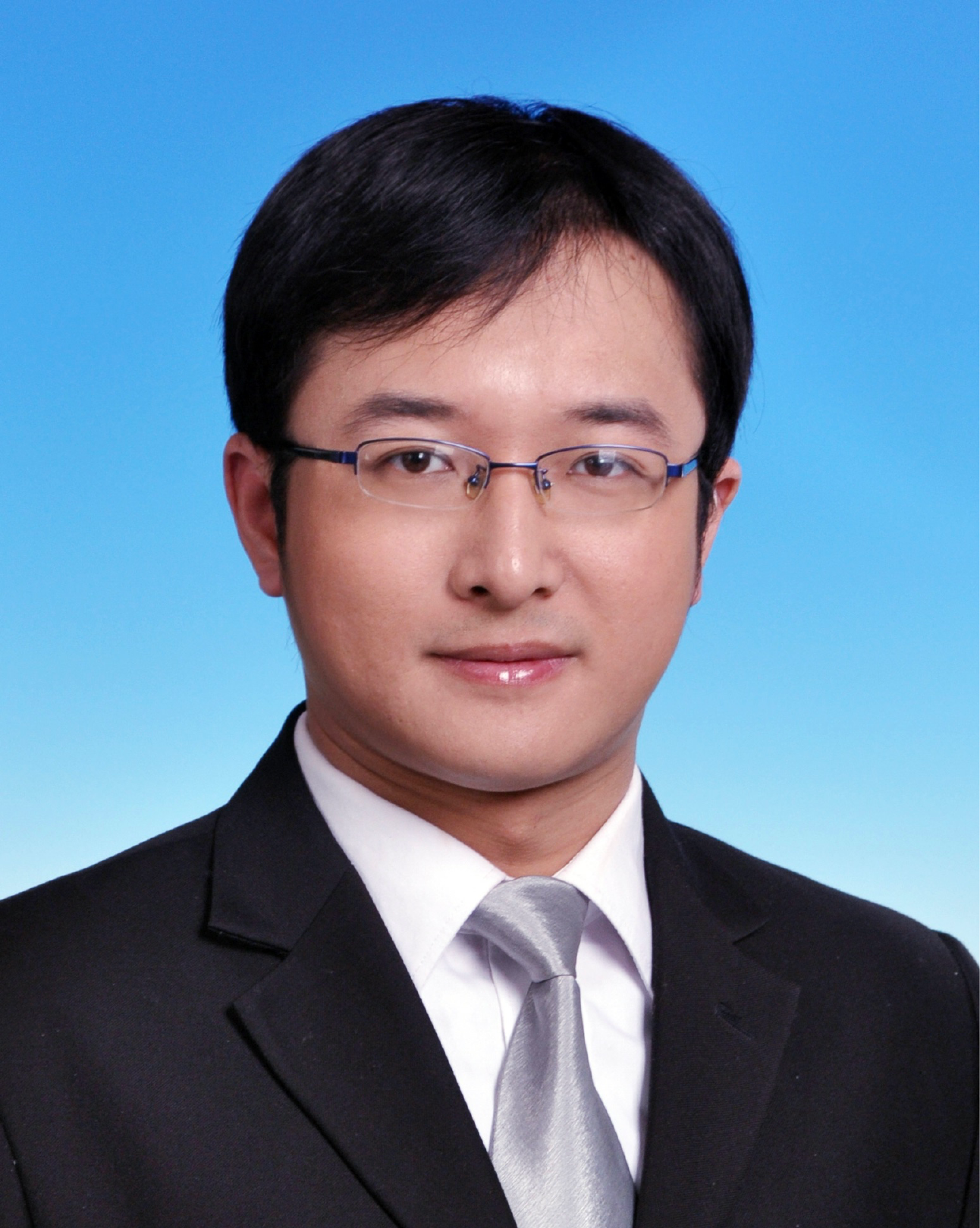}}]{Ke Tang}
(Senior Member, IEEE) received the B.Eng. degree from the Huazhong University of Science and Technology, Wuhan, China, in 2002, and the Ph.D. degree from Nanyang Technological University, Singapore, in 2007.
He is currently a Professor with the Department of Computer Science and Engineering, Southern University of Science and Technology (SUSTech), Shenzhen, China. Before joining SUSTech in January 2018, he was with the School of Computer Science and Technology, University of Science and Technology of China, Hefei, China, first as an Associate Professor from 2007 to 2011, and then as a Professor from 2011 to 2017. His major research interests include evolutionary computation and machine learning, particularly in large-scale evolutionary computation, integration of evolutionary computation, and machine learning, as well as their applications. He has published more than 180 papers, which have received over 12000 Google Scholar citations with an H-index of 54. 
Prof. Tang received the IEEE Computational Intelligence Society Outstanding Early Career Award in 2018, the Newton Advanced Fellowship (Royal Society) in 2015, and the Natural Science Award of Ministry of Education of China in 2011 and 2017. He is a Changjiang Scholar Professor (awarded by the MOE of China). He is an Associate Editor of the IEEE TRANSACTIONS ON EVOLUTIONARY COMPUTATION and served as a member on the editorial boards for a few other journals.
\end{IEEEbiography}

\vspace{11pt}

\vfill

\end{document}